\documentclass[10pt,twocolumn,letterpaper]{article}

\usepackage[pagenumbers]{cvpr} %
\usepackage{booktabs}
\usepackage{multirow}

\usepackage{pifont}
\newcommand{\cmark}{\ding{51}}
\newcommand{\xmark}{\ding{55}}

\makeatletter
\def\blfootnote{\gdef\@thefnmark{}\@footnotetext}
\makeatother

\definecolor{cvprblue}{rgb}{0.21,0.49,0.74}
\usepackage[pagebackref,breaklinks,colorlinks,allcolors=cvprblue]{hyperref}

\def\paperID{108} %
\def\confName{3DV\xspace}
\def\confYear{2027\xspace}

\newcommand{\methodname}{SplashSplat\xspace}
\title{\methodname: Reconstructing Splashing Liquids from Real-World \\ Multi-View Videos}

\author{Peiyu Liu$^{1*}$ \quad Dingxi Zhang$^{2*}$ \quad Federico Tombari$^{3}$ \quad Marc Pollefeys$^{2,4}$ \\ 
Christina Tsalicoglou$^{3}$ \quad Daniel Barath$^{2}$\\
$^{1}$ EPFL \quad
$^{2}$ ETH \quad
$^{3}$ Google \quad
$^{4}$ Microsoft
}

\selectfont

\begin{document}

\twocolumn[{%
	\renewcommand\twocolumn[1][]{#1}%
	\maketitle
    \vspace{-2em}
	\includegraphics[width=\linewidth]{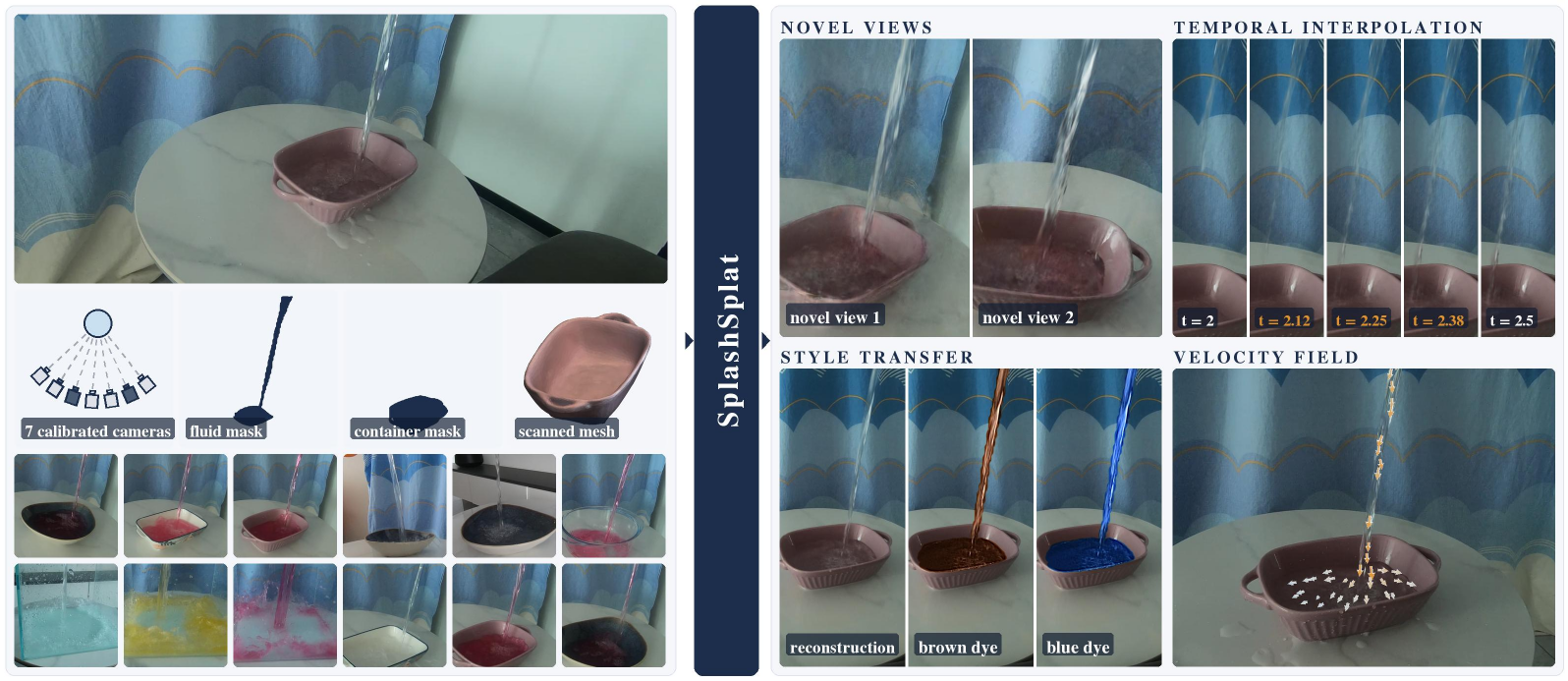}
    \captionof{figure}{\textbf{\methodname{} reconstructs splashing
liquids from real multi-view video.} \textbf{Left:} our benchmark
captures pouring and splashing across diverse containers, liquids,
and backgrounds with seven calibrated cameras, with fluid and
container masks and scanned container meshes. \textbf{Right:} from
these observations, \methodname{} synthesizes novel views, recovers a
velocity field, and supports temporal interpolation and style
transfer.}
	\label{fig:teaser}
}]
\blfootnote{$^*$ Equal Contribution. \quad Email: {\tt\small peiyu.liu@epfl.ch}}

\begin{abstract}
A splash lives for a fraction of a second: sheets tear into ligaments
and droplets, appearance is view-dependent and nearly textureless,
and little persists long enough to track. Reconstruction research has
consequently focused on smoke, synthetic liquids, or gently deforming
surfaces. To our knowledge, no synchronized multi-view dataset of
splashing liquids exists. We therefore introduce a benchmark of 20 real scenes, from coherent streams to violent splashes, captured by seven synchronized, calibrated 4K cameras at 60\,fps, with manually refined per-view liquid and container masks and fixed evaluation splits. We further present \methodname{}, built on a single principle:
\emph{impose physical structure only where the observations can
constrain it}. Per-frame liquid SDFs fused from the masks provide the
geometry, level-set transport between consecutive SDFs yields a
coarse velocity field, and Lagrangian carriers advected along this
flow, corrected against each new observation and reseeded where
coverage is lost, decode local Gaussians for differentiable rendering. \methodname{} outperforms state-of-the-art dynamic Gaussian splatting methods on our real captures and on a synthetic benchmark, with physically more plausible motion and a lower training cost. The same representation supports temporal interpolation and style transfer without re-optimization. Project Page: \url{https://niko-creater.github.io/splashsplat-web/}
\end{abstract}
    
\vspace{-2em}
\section{Introduction}
\label{sec:intro}

A falling stream shatters into sheets, the sheets tear into ligaments, the ligaments into
droplets, and an instant later the surface is calm again. Almost
nothing in the scene persists long enough to be tracked. Capturing
such events digitally has long been a goal spanning computer
graphics, vision, and experimental fluid
mechanics~\cite{elsinga2006tomographic, eckert2019scalarflow}: a
faithful replica of a real splash would benefit visual effects,
provide reference data for validating fluid simulators, and supply
dynamic assets for embodied AI. Yet, while 3D Gaussian
Splatting~\cite{kerbl20233d} has made photorealistic reconstruction
of general dynamic scenes practical~\cite{luiten2024dynamic,
wu20244d, Li_STG_2024_CVPR}, splashing liquids have remained out of
reach.

This gap has two causes. The first is intrinsic difficulty. Liquid
appearance is view-dependent and weakly textured, undermining
correspondence. Thin sheets, ligaments, and droplets emerge and
vanish within a few frames, defeating representations that assume
persistent geometry. And the boundary conditions a simulator would
need, such as inflow rate and total volume, cannot be observed from
images. Existing approaches are strained from both directions.
Classic capture systems recovered gently deforming water with
specialized apparatus~\cite{ihrke2005reconstructing,
morris2011dynamic, wang2009physically}, and modern fluid
reconstruction methods model the medium as a semi-transparent density
field whose interior is optically observable, an assumption that
holds for smoke~\cite{chu2022physics, yu2023inferring,
gao2025fluidnexus, xie2025fluidgs} but not for an opaque liquid
surface. General dynamic 3DGS methods~\cite{yang2023deformable3dgs,
Li_STG_2024_CVPR, cho20264d, lu2024scaffold} make no such assumption,
but grant every primitive independent motion, far more freedom than a
few views of near-textureless liquid can constrain, and in practice
let primitives fade and reappear rather than move with the
liquid~\cite{xing2022differentiable, gao2026surfphase}.

The second cause is the absence of data. Real liquid captures are
limited to gently deforming surfaces, static
states~\cite{wang2024towards}, or two-view observations of
boiling~\cite{gao2026surfphase}, so no benchmark has been able to
measure how existing methods behave on liquids in motion. We address
this with the first synchronized multi-view video benchmark of
splashing liquids: 20 scenes recorded by seven
calibrated cameras at 4K and 60\,fps, spanning coherent streams to
violent splashing across a range of containers, liquids, and
backgrounds, each with manually refined per-view liquid and container
masks and a scanned container mesh.

We further present \methodname{}, built on a simple principle:
\emph{impose physical structure only where observations can constrain
it}. Without the boundary conditions that images do not reveal, a fluid
solver produces motion that is confident but unverifiable. We instead retain only
the \emph{kinematics} of fluids, level-set transport of the
interface~\cite{enright2002animation, wang2009physically},
approximate incompressibility, and continuum deformation, and let
observations close the loop. Concretely, \methodname{} fuses the
per-frame multi-view masks into liquid SDFs, fits a coarse velocity
field between consecutive SDFs, and advects Lagrangian carriers along
this flow. Each carrier is corrected against the next observation and reseeded
where coverage is lost, then decodes a small set of local Gaussians.
Motion is thus estimated at the granularity the data supports, while
fine geometry and view-dependent appearance are recovered by
differentiable rendering.

We evaluate \methodname{} against state-of-the-art dynamic Gaussian
splatting methods~\cite{yang2023deformable3dgs, Li_STG_2024_CVPR,
cho20264d} on our real captures and on the synthetic NeuroFluid
benchmark~\cite{guan2022neurofluid}. Across the baselines, rendering
quality and physical plausibility trade off against each other, the
expected consequence of recovering motion from appearance alone on a
weakly textured surface: a fit that explains the training views need
not correspond to how the liquid actually moved. \methodname{} improves both, with the lowest training time and peak memory of all methods compared.
Our contributions are summarized as follows:
\begin{itemize}
    \item We introduce, to our knowledge, the first synchronized
    multi-view video benchmark of splashing liquids: 20 scenes
    captured with a calibrated seven-camera rig, annotated with
    refined per-view liquid and container masks and scanned container
    meshes.
    \item We present \methodname{}, which imposes physical structure
    only where observations can constrain it: fluid kinematics
    transports Lagrangian carriers between frames, and each new
    observation corrects them, avoiding the unobservable inflow and
    volume conditions a simulator would require.
    \item We show that this design outperforms deformation-based
baselines and a full fluid solver in reconstruction quality and
physical plausibility, at lower training cost, and yields a
representation that supports temporal interpolation and style
transfer.
\end{itemize}

\begin{table*}[t]
\centering
\caption{\textbf{Comparison with existing fluid datasets.} Real-world
fluid data is dominated by smoke, while real liquid data is confined
to static states or to interfacial dynamics from at most two
viewpoints. Ours is the only synchronized, metrically calibrated
multi-view video benchmark of splashing liquids. Sync.: synchronized
multi-camera capture. Masks: per-view liquid masks.
$^\ddagger$Rendered virtual cameras rather than a physical rig.
$^\dagger$A single hand-held camera moved around static scenes.
--: not applicable or not reported.}
\label{tab:dataset_comparison}
\small
\setlength{\tabcolsep}{4.5pt}
\begin{tabular}{l ll r c c l r l c}
\toprule
Dataset & Source & Medium & Scenes & Views & Sync. & Resolution & FPS & Dynamics & Masks \\
\midrule
ScalarFlow~\cite{eckert2019scalarflow}  & Real    & Smoke  & 104          & 5 & \cmark & $1080\times1920$ & 60 & rising plume            & \xmark \\
FluidNexus~\cite{gao2025fluidnexus}     & Real    & Smoke  & 240 & 5 & \cmark & $1080\times1920$ & 30 & jet, solid interaction  & \xmark \\
NeuroFluid~\cite{guan2022neurofluid}    & Synth.\ & Liquid & 3            & 5$^\ddagger$ & -- & $400\times400$   & -- & simulated free surface     & \xmark \\
TransProteus~\cite{eppel2022predicting} & Synth.\ & Liquid & 50k imgs.\   & 1$^\ddagger$ & -- & --               & -- & liquid in vessels          & \cmark \\
Phys-Liquid~\cite{ma2026phys}           & Synth.\ & Liquid & 5            & 6$^\ddagger$ & -- & $1280\times720$               & -- & container-induced sloshing & \cmark \\
DTLD~\cite{wang2024towards}             & Real    & Liquid & 3            & 1$^\dagger$  & \xmark & $1280\times720$ & 30 & static levels           & \cmark \\
SurfPhase~\cite{gao2026surfphase}       & Real    & Liquid & 1            & 2            & \cmark & --             & 2000 & pool boiling             & \xmark \\
\midrule
\textbf{\methodname{} (Ours)}           & Real    & Liquid & 20           & \textbf{7}   & \cmark & $\mathbf{3840\times2160}$ & 60 & \textbf{splashing, break-up} & \cmark \\
\bottomrule
\end{tabular}
\end{table*}

\section{Related Work}
\label{sec:related}

\vspace{1mm}\noindent\textbf{Dynamic 3D Reconstruction.}
Dynamic neural rendering extends static reconstruction to
time-varying scenes, initially via canonical-space deformation or
motion fields~\cite{pumarola2021d, park2021nerfies, li2021neural,
park2021hypernerf}. Gaussian-based representations render
substantially faster: Dynamic 3D Gaussians~\cite{luiten2024dynamic}
tracks persistent primitives, deformation-based
methods~\cite{yang2023deformable3dgs, wu20244d} warp canonical
Gaussians through learned fields, others model native 4D
spatiotemporal primitives~\cite{yang2024real} or equip primitives
with temporal opacity and polynomial motion
trajectories~\cite{Li_STG_2024_CVPR}, and anchor-based variants
decode local Gaussians from a sparse scaffold~\cite{lu2024scaffold,
cho20264d}, a scheme our representation builds upon. These methods
achieve high-quality novel-view synthesis on textured scenes, but
recover motion from photometric gradients, which are ineffective when
appearance provides little spatial signal: primitives tend to remain
in place and adjust their scale and color to match the target rather
than translate with the surface~\cite{xing2022differentiable,
gao2026surfphase}. A near-textureless liquid is exactly this regime,
and our carriers instead follow a flow fitted to the observed
interface.

\vspace{1mm}\noindent\textbf{Fluid Reconstruction from Video.}
Capturing real liquids predates neural rendering: fluorescent
labeling~\cite{ihrke2005reconstructing}, refraction
stereo~\cite{morris2011dynamic}, camera
arrays~\cite{ding2011dynamic}, angular-domain surface
recovery~\cite{ye2012angular}, appearance transfer from sparse
views~\cite{okabe2015fluid}, and physically guided stereoscopic
modeling~\cite{wang2009physically} recovered water geometry with
specialized apparatus. These systems address gently deforming
surfaces, producing geometry without appearance or novel-view
synthesis. Modern approaches reconstruct fluid states from video
under physical constraints, from tomographic and transport-based
formulations~\cite{eckert2019scalarflow, franz2021global,
gregson2014capture, atcheson2008time} to physics-informed neural and
Gaussian methods~\cite{chu2022physics, yu2023inferring,
xie2025fluidgs, wang2024pict, tao2025flowcapx, ni2025dfk,
zhang2026gausmoke, gao2025fluidnexus, tao2026lagrangiansplats}. All
are developed and evaluated on smoke, a participating medium with
diffuse boundaries. An opaque liquid instead terminates rays at a sharp,
topology-changing interface: the interior is never observed, so the
unknown is a free surface rather than a volumetric
field~\cite{gao2026surfphase}. This is what leads us to represent the
liquid by an observed SDF and surface-bound carriers rather than by a
density field.

For liquids, NeuroFluid~\cite{guan2022neurofluid} and
GaussFluids~\cite{du2025gaussfluids} couple particle transition
models or transported Gaussians with differentiable rendering, a
formulation close to ours, but evaluate on synthetic DFSPH/Blender
sequences whose liquids are rendered transparent.
SurfPhase~\cite{gao2026surfphase} works with real liquid, but on the
interfacial dynamics of boiling from two views, compensating for the
sparse viewpoints with a video diffusion prior rather than with
physical structure. Contained-liquid geometry is also estimated from
single images for perception and
manipulation~\cite{eppel2022predicting, ma2026phys,
richter2022image}, on quasi-static laboratory liquids, while
PhysGaussian~\cite{xie2024physgaussian} and Gaussian
Splashing~\cite{feng2025gaussian} couple simulation with Gaussians to
synthesize motion in already-reconstructed scenes rather than to
reconstruct it. Reconstructing free-surface liquids undergoing
splashing and break-up from real multi-view video therefore remains
open.

\vspace{1mm}\noindent\textbf{Fluid Datasets.}
Real-world fluid benchmarks are dominated by smoke:
ScalarFlow~\cite{eckert2019scalarflow} provides five-view captures of
buoyant plumes, and FluidNexus~\cite{gao2025fluidnexus} adds two
120-scene five-view smoke datasets with textured backgrounds, one of
them featuring fluid--solid interaction. Liquid data is either
synthetic (NeuroFluid~\cite{guan2022neurofluid},
TransProteus~\cite{eppel2022predicting}, and
Phys-Liquid~\cite{ma2026phys} render simulated liquids) or restricted
in dynamics and viewpoints: DTLD~\cite{wang2024towards} records
static liquid states scanned by a hand-held moving camera, laboratory
captures target segmentation from static single
views~\cite{narasimhan2022self}, and
SurfPhase~\cite{gao2026surfphase} provides a single synchronized
dual-view boiling sequence alongside uncalibrated monocular footage.
As \cref{tab:dataset_comparison} summarizes, no existing dataset
provides synchronized multi-view video of splashing liquids, which is
what our benchmark contributes.

\section{The \methodname{} Benchmark}
\label{sec:dataset}

\begin{figure*}[t]
    \centering
    \includegraphics[width=0.98\linewidth]{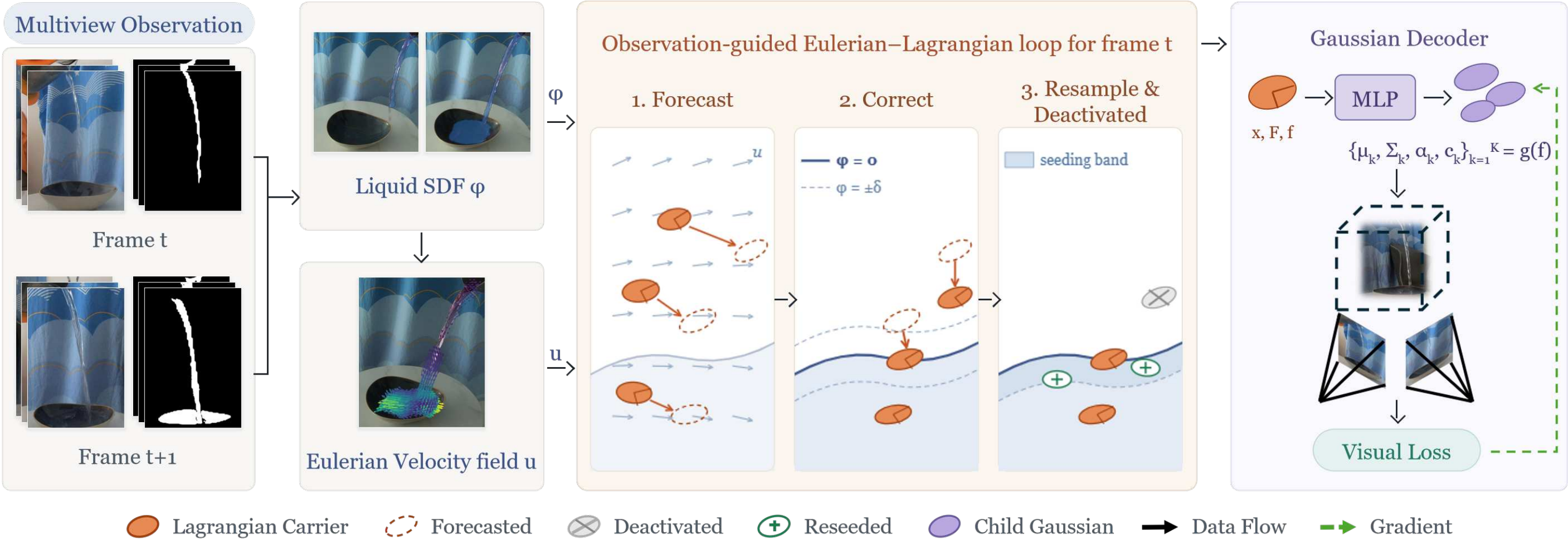}
    \caption{\textbf{Overview of \methodname{}.} At each frame, per-view
liquid masks are fused into an observed liquid SDF, while the static
scene is represented by pretrained background Gaussians
(\cref{sec:obs_fields}). Consecutive SDFs yield a coarse velocity
field that advects Lagrangian carriers through a forecast--correct--resample loop: carriers are propagated by the
flow, corrected against the next-frame SDF to suppress exterior
drift, and resampled to restore missing surface coverage
(\cref{sec:carriers}). Each carrier decodes local Gaussians, which
are rasterized together with the frozen background and optimized
under photometric and mask supervision
(\cref{sec:gaussian_reconstruction,sec:optimization}).}
    \label{fig:pipeline}
\end{figure*}

Splashing liquids are unlike the smoke that existing physics-informed
benchmarks target: reflection and refraction make correspondence
unreliable, and thin structures form and break up within a few
frames. No existing capture records these dynamics in synchronized
multi-view video (\cref{tab:dataset_comparison}). We therefore
introduce a benchmark for novel-view reconstruction of splashing
liquids (\cref{fig:teaser}).

\vspace{1mm}\noindent\textbf{Capture and calibration.}
Each scene is recorded by seven GoPro cameras at $3840\times2160$ and
60\,fps, distributed along a sparse arc that subtends $77^\circ$ on
average about the scene center. The cameras are free-running, so we
recover a common timeline post hoc by cross-correlating per-camera
audio envelopes; the residual synchronization error is bounded by
half a frame interval ($8.3$\,ms, RMS $4.8$\,ms). Intrinsics are
calibrated per camera with a ChArUco board. Estimating relative poses
is harder, as the sparse viewpoints share limited feature overlap: we therefore record a texture-rich static scene to provide sufficient visual features for robust COLMAP-based camera registration, additionally sweeping a handheld camera across the setup to bridge otherwise weakly connected fixed viewpoints. Bundle adjustment converges to a mean reprojection error
of $0.9$\,px, and the cameras remain fixed for the session.

\vspace{1mm}\noindent\textbf{Scenes.}
The benchmark contains 20 scenes spanning a range of pouring
dynamics, from thin coherent streams impacting a shallow container to
larger volumes released into a deep tank, which produce violent
splashes, thin sheets, detached droplets, and frequent topology
changes. Scenes vary in container shape and size, liquid color, and
background; transparent containers additionally introduce
view-dependent refraction, making those scenes substantially harder
for multi-view reconstruction. For each scene we identify the
interval of effective liquid motion and uniformly sample 50
synchronized frames from it, giving $7{,}000$ images that cover the
main phases of the deformation at a consistent sampling density.

\vspace{1mm}\noindent\textbf{Annotations and protocol.}
Per-view liquid and container masks are initialized with
SAM3~\cite{carion2026sam} and manually refined; refinement is almost
entirely additive, since SAM3 under-segments thin streams and water
seen through glass. We additionally reconstruct a mesh of each
container using~\cite{arcode2024}, providing the static
geometry the liquid interacts with. The benchmark defines a fixed
split: five cameras for training and two test cameras,
$8^\circ$--$15^\circ$ from their nearest training view, reserved for
novel-view evaluation. All observation quantities consumed by any
method, including the fused SDFs of \cref{sec:method}, are computed
from training views only. Full calibration
and annotation details are in the supplementary.

\section{Method}
\label{sec:method}

Given the synchronized, calibrated images $\mathcal{I}=\{I_{t,v}\}$
and liquid masks $\mathcal{M}=\{M_{t,v}\}$, we
reconstruct a dynamic foreground $\mathcal{G}^{\mathrm{fg}}_t$ and
render it with a static background $\mathcal{G}^{\mathrm{bg}}$,
\begin{equation}
    \hat I_{t,v}
    =
    \mathcal{R}\left(
    \mathcal{G}^{\mathrm{bg}} \cup \mathcal{G}^{\mathrm{fg}}_t;\;
    \Pi_v\right),
    \label{eq:rendering}
\end{equation}
where $\mathcal{R}$ is differentiable Gaussian
rasterization~\cite{kerbl20233d} and $\Pi_v$ the calibrated camera
$v$. Our design follows a single principle: \emph{impose physical
structure only where the observations can constrain it}. A momentum-based solver requires pressure and boundary conditions
that are not observable from real-world captures, and when
misspecified it produces motion that is self-consistent yet incompatible with the observed liquid (\cref{sec:ablation}). We
therefore retain only fluid \emph{kinematics}: observed geometry,
interface transport under approximate incompressibility, and
continuum deformation. We recover observation fields from the masks
(\cref{sec:obs_fields}), advect Lagrangian carriers through a
forecast--correct--resample loop (\cref{sec:carriers}), and decode
them into Gaussians optimized by rendering
(\cref{sec:gaussian_reconstruction,sec:optimization}). The overview of our pipeline is shown in \cref{fig:pipeline}.

\subsection{Observation Fields from Multi-View Masks}
\label{sec:obs_fields}

\noindent\textbf{Observed geometry.}
For each frame $t$, we fuse the calibrated training-view masks into a
3D occupancy by visual-hull carving on an isotropic voxel grid
and compute its signed distance field (SDF)
$\phi_t:\mathbb{R}^3\rightarrow\mathbb{R}$, with $\phi_t<0$ inside the
observed liquid. The hull is asymmetric evidence: a point outside it
is contradicted by at least one view, whereas the interior remains
uncertain, since concavities that no view can carve are retained.
This determines where the observations may overrule the motion model.

\vspace{1mm}\noindent\textbf{Observed motion.}
Motion is far less constrained by the observations: liquid appearance
is weakly textured and view-dependent, and motion within the volume
is not observable. The evolving interface nonetheless provides a kinematic constraint, since it approximately satisfies the level-set equation $\partial_t\phi + u\cdot\nabla\phi = 0$ under approximate incompressibility, $\nabla\cdot u \approx 0$. We
therefore fit a coarse velocity field $u_t$ to this transport
relation between consecutive
observations~\cite{wang2009physically},
\begin{equation}
    u_t = \operatorname*{arg\,min}_{u}
    \sum_{q\in\mathcal{N}_t}
    \Big(\phi_{t+1}(q)-\phi_t\big(q-\Delta t\,u(q)\big)\Big)^2,
    \label{eq:velocity_fit}
\end{equation}
where $u$ is defined on a coarse grid and $\mathcal{N}_t$ is a narrow
band around the two interfaces. The objective is minimized
iteratively, smoothing each iterate and projecting it toward its
divergence-free component, which propagates the interface constraint
into the interior, under a bound on velocity magnitude. Since
level-set transport is insensitive to motion tangential to the
interface, \cref{eq:velocity_fit} does not
recover the full flow; $u_t$ therefore serves as a short-range
transport prior, with the residual left to the per-frame observations.

\subsection{Forecast, Correct, Resample}
\label{sec:carriers}

A motion estimate is useful for propagating information between
adjacent observations, but too uncertain to determine the liquid
geometry in an open loop. We therefore combine the Eulerian fields
above with advected Lagrangian carriers: the fields provide per-frame
geometry and a one-step motion proposal, while the carriers convert
this proposal into positions and local deformations that drive the
Gaussian representation. The carrier set is
\begin{equation}
    \mathcal{P}
    =
    \left\{\left(b_i, e_i,
    \{x_{i,t}, F_{i,t}\}_{t=b_i}^{e_i}\right)\right\}_{i=1}^{N_p},
    \label{eq:carrier_union}
\end{equation}
where $x_{i,t}\in\mathbb{R}^3$ is position,
$F_{i,t}\in\mathbb{R}^{3\times3}$ a local deformation gradient, and
$b_i, e_i$ the birth and end frames; only the active set
$\mathcal{A}_t=\{i \mid b_i\leq t\leq e_i\}$ is decoded at time $t$.
Finite lifetimes let the representation follow inflow, break-up, and
topology change without requiring one particle set to remain valid
over the entire sequence.

\vspace{1mm}\noindent\textbf{Forecast.}
Each active carrier is propagated with midpoint RK2, and its
deformation gradient updated from the local velocity gradient:
\begin{align}
    x_{i,t+\frac{1}{2}}
    &= x_{i,t} + \tfrac{\Delta t}{2}\, u_t(x_{i,t}),
    \nonumber\\
    x^{-}_{i,t+1}
    &= x_{i,t} + \Delta t\, u_t(x_{i,t+\frac{1}{2}}),
    \nonumber\\
    F^{-}_{i,t+1}
    &= \left(I + \Delta t\, \nabla u_t(x_{i,t})\right) F_{i,t},
    \label{eq:carrier_forecast}
\end{align}
with the singular values of $F$ bounded to avoid degenerate
stretching. The field $u_t$ transports carrier centers, while its
gradient evolves $F$ by the standard continuum relation
$\dot F = \nabla u\,F$, so each carrier records how the flow has
locally rotated and stretched its neighborhood. This frame later
orients the decoded Gaussians, which is how the transported motion
reaches the rendered image.

\vspace{1mm}\noindent\textbf{Correction.}
Coarse transport drifts from the observed surface. Since the fused
SDF is a visual hull, its evidence is one-sided: a carrier predicted
outside the hull is contradicted by at least one view, while one
predicted inside is not. We therefore correct exterior predictions
and leave interior carriers untouched. With
$\phi^{-}_i=\phi_{t+1}(x^{-}_{i,t+1})$ and unit normal $\hat n_i$ at
that point,
\begin{equation}
    x_{i,t+1}
    =
    x^{-}_{i,t+1}
    -
    \mathbf{1}_{\{\phi^{-}_i>0\}}\,
    \min(\phi^{-}_i, \delta)\, \hat n_i,
    \label{eq:sdf_correction}
\end{equation}
which is not written back into $u_t$. Carriers remaining outside the
observed liquid beyond a distance threshold are deactivated.

\vspace{1mm}\noindent\textbf{Resampling.}
Correction moves carriers but does not create them, while liquid
enters the scene continuously. We therefore restore coverage from the
current observation: the near-surface band is discretized into cells, and any
cell whose active carrier count falls below a target $\bar n$
receives new carriers, inheriting the local deformation of nearby
active carriers when available. The three steps close a cycle in
which the fitted field propagates motion, the carriers hold local
temporal structure, and each new observation removes drift and
restores coverage.

\subsection{Decoding Carriers into Gaussians}
\label{sec:gaussian_reconstruction}

Since the liquid interior is unobservable, we spend representational
capacity near the interface: carriers live in a near-surface band and
their Gaussians are suppressed away from it, making the foreground a
thin deforming shell. Each active carrier $i$ carries a learnable
feature $f_i$ and base scale $s_i$, which together with its
propagated deformation gradient define a local frame
$A_{i,t}=F_{i,t}\operatorname{diag}(s_i)$. Following anchor-based
decoding~\cite{lu2024scaffold, cho20264d}, a set of heads shared
across all carriers and frames maps $f_i$ to $K$ child Gaussians
expressed in this frame. The offset head $g_o$ gives a bounded
residual $o_{i,k}=\beta\tanh(g_o(f_i)_k)$, and the scale head $g_s$ a
scale residual, yielding
\begin{equation}
    \mu_{i,k,t}=x_{i,t}+A_{i,t}\,o_{i,k},
    \,
    B_{i,k,t}
    =
    \tfrac{1}{\sqrt K}A_{i,t}
    \operatorname{diag}\!\big(\exp(\eta_s g_s(f_i)_k)\big),
    \label{eq:child_position}
\end{equation}
with $\Sigma_{i,k,t}=B_{i,k,t}B_{i,k,t}^{\top}$. Both the offset and
the covariance are expressed in $A_{i,t}$, so the children remain
bounded within their carrier and inherit its orientation and
stretch. Opacity is modulated by the observed SDF,
\begin{equation}
    \alpha_{i,k,t}
    =
    \sigma\big(g_\alpha(f_i)_k\big)\,
    \sigma\big(\kappa_\alpha(\tau_\alpha-|\phi_t(x_{i,t})|)\big),
    \label{eq:sdf_opacity}
\end{equation}
suppressing Gaussians far from the interface. Geometry is view
independent, while colour is conditioned on the viewing direction
$\omega_{i,t,v}$ through $g_c([f_i,\omega_{i,t,v}])$, letting the
children absorb specular highlights and the background seen through
the liquid without modelling refraction explicitly.

\subsection{Optimization}
\label{sec:optimization}

The static background $\mathcal{G}^{\mathrm{bg}}$ is pretrained per
scene with Scaffold-GS~\cite{lu2024scaffold} and kept frozen. At each
iteration we sample a time--camera pair $(t,v)$ and rasterize the
active foreground Gaussians together with the background. Since the
frozen background cannot absorb residuals on background pixels, a
whole-image loss would push the foreground to explain them. We
therefore supervise the liquid region only, normalizing by the masked
pixel count so that thin streams and large splashes contribute
equally. On top of this photometric term we add a gradient term under
the same normalization, which sharpens thin structures; a D-SSIM
term~\cite{kerbl20233d} on the mask bounding box; a silhouette term
matching rendered foreground opacity to $M_{t,v}$; and a regularizer
on Gaussian covariance. Two terms follow~\cite{gao2025fluidnexus}:
$\mathcal{L}_{\mathrm{sep}}$ keeps the $K$ children of a carrier from
collapsing onto one another, and $\mathcal{L}_{\mathrm{col}}$
penalizes colour variance among children whose carriers share a grid
cell, suppressing flicker as carriers are replaced. The objective is
their weighted sum. Definitions and weights are in the supplementary
material.

\begin{table*}[ht]
\centering
\caption{\textbf{Quantitative comparison on our real-world
benchmark.} Foreground novel-view synthesis, temporal consistency,
physical plausibility, and efficiency. All methods use the same protocol on native-resolution fluid-centric crops, with training time on a single NVIDIA V100; ours includes SDF
preprocessing and both foreground and background reconstruction.}
\label{tab:nvs_fg}
\small
\setlength{\tabcolsep}{3pt}
\begin{tabular}{l cccc cc ccc}
\toprule
& \multicolumn{4}{c}{Novel View Synthesis}
& \multicolumn{2}{c}{Physical Plausibility}
& \multicolumn{2}{c}{Efficiency} \\
\cmidrule(lr){2-5}
\cmidrule(lr){6-7}
\cmidrule(lr){8-10}

Method
& PSNR$\uparrow$
& SSIM$\uparrow$
& LPIPS$\downarrow$
& $|\Delta\mathrm{Jitter}|\downarrow$
& $\sigma_D\!\downarrow$
& $\Delta E\!\downarrow$
& Train Time$\downarrow$
& Memory$\downarrow$
&  \\

\midrule
Deformable-3DGS~\cite{yang2023deformable3dgs}
& 12.95
& 0.9243
& 0.1307
& 0.0501
& 529.4
& 0.426
& 194.2\,min
& 17.53\,GiB \\

SpacetimeGaussians~\cite{Li_STG_2024_CVPR}
& 15.78
& 0.9442
& 0.1088
& 0.0504
& 878.8
& 0.478
& 68.8\,min
& 6.58\,GiB \\

4D-Scaffold-GS~\cite{cho20264d}
& 18.92
& 0.9552
& 0.0898
& 0.0428
& 1005.1
& 0.436
& 42.1\,min
& 16.78\,GiB \\

\midrule
\methodname{} (Ours)
& \textbf{20.13}
& \textbf{0.9570}
& \textbf{0.0806}
& \textbf{0.0276}
& \textbf{219.9}
& \textbf{0.373}
& \textbf{37.8\,min}
& \textbf{5.31\,GiB} \\

\bottomrule
\end{tabular}
\end{table*}

\section{Experiments}
\label{sec:exp}

\begin{figure*}[t]
  \centering
  \setlength{\tabcolsep}{1pt}
  \renewcommand{\arraystretch}{0.85}
  \newcommand{\shw}{0.094\linewidth}
  \newcommand{\cs}{figures/crop_showcase_grid}

  \includegraphics{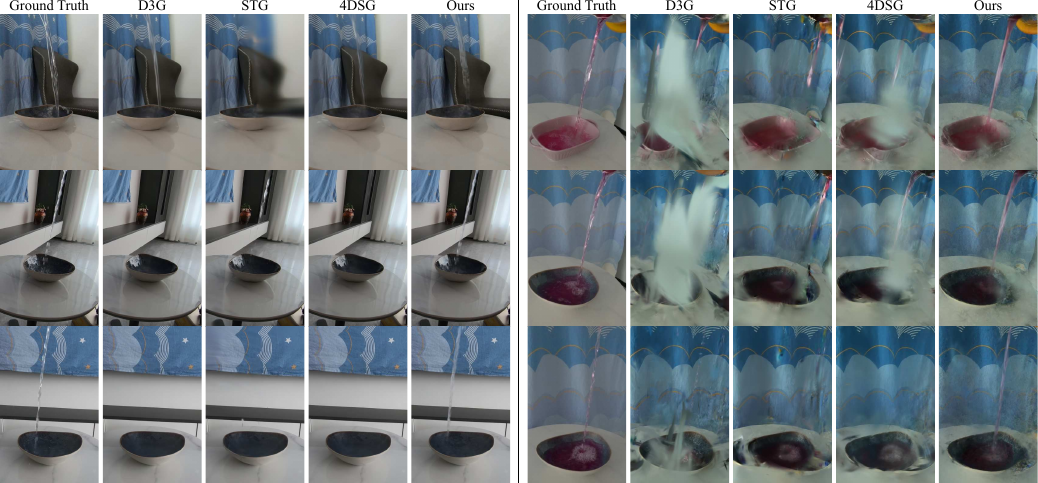}

  \caption{\textbf{Qualitative comparison on real captures.} Training views (left) and test views (right), one scene per row. On test views the baselines place liquid Gaussians away from the observed surface: D3G collapses into a full-frame plume, while STG and 4DSG leave haze that occludes the container. \methodname{} keeps the stream continuous and the liquid confined to the observed surface.}
  \label{fig:crop_showcase}
\end{figure*}

\paragraph{Datasets.}
Our primary evaluation is on our real multi-view captures of
splashing liquids (\cref{sec:dataset}). To further verify
reconstruction quality, we also evaluate on the widely-adopted
synthetic NeuroFluid dataset~\cite{guan2022neurofluid}, generated
with DFSPH~\cite{bender2015divergence} and rendered in Blender from
5 frontal cameras.

\vspace{1mm}\noindent\textbf{Baselines.}
We compare against three dynamic Gaussian splatting methods: Deformable-3DGS (D3G)~\cite{yang2023deformable3dgs}, SpacetimeGaussians (STG)~\cite{Li_STG_2024_CVPR}, and 4D-Scaffold-GS (4DSG)~\cite{cho20264d}. All methods are trained on the same views with their released implementations and recommended settings.
Fluid-specific methods are either inapplicable, since a volumetric density model cannot represent an opaque free surface (\cref{sec:related}), or not open-sourced~\cite{du2025gaussfluids}.

\vspace{1mm}\noindent\textbf{Metrics.}
To evaluate novel view synthesis quality, we report PSNR, SSIM~\cite{wang2004ssim}, and LPIPS~\cite{lpips} on test views. Temporal consistency is measured by $|\Delta\mathrm{Jitter}|$, the absolute difference between the jitter of a rendered sequence and that of the ground-truth video. Jitter~\cite{du2025gaussfluids} is defined as the standard deviation of per-pixel intensity differences between adjacent frames, pooled over fluid-mask pixels. We report the jitter error rather than the raw jitter, since a static or overly smoothed foreground can trivially achieve near-zero jitter without matching the true temporal dynamics. To quantify physical fidelity, we follow GaussFluids~\cite{du2025gaussfluids} by tracking two key metrics on foreground fluid: the per-frame standard deviation of SPH density estimates $\sigma_D$ to evaluate incompressibility, and the deviation of mechanical energy from monotone dissipation $\Delta E$ to assess plausible dissipation dynamics. 
Finally, we report computational cost: training time and peak memory.

\vspace{1mm}\noindent\textbf{Implementation details.}
For each scene, we carve the observed liquid hull on a near-isotropic
grid with $176$ cells along its longest axis and fit the velocity
field on a coarser $96$-cell grid within a narrow band around the
interface. 
Each carrier holds a $32$-dimensional feature decoded by three linear heads
(offset, scale, opacity) and a two-layer colour MLP of width $64$
conditioned on the viewing direction, producing $K=2$ Gaussian
children. Since the liquid occupies only 5--8\% of the raw image
area, we crop each view around the liquid region using the bounding
box of the union of multi-view foreground masks before training,
giving an average input resolution of $1200\times1800$. We first
optimize the static background for $20$k iterations and
keep it fixed thereafter, while the dynamic foreground is optimized
for $3$k iterations at the cropped resolution using
Adam~\cite{kingma2014adam} with learning rate $2\times10^{-3}$.
All experiments run on a single NVIDIA V100-32GB GPU. Further details are provided in the
supplementary material.

\subsection{Comparison to Baselines}
\label{sec:real_results}

\begin{figure}[t]
  \centering
  \setlength{\tabcolsep}{1pt}
  \renewcommand{\arraystretch}{0.98}
  \newcommand{\shw}{0.176\linewidth}
  \newcommand{\wsdir}{figures/watersphere_showcase/watersphere_5}
  \begin{tabular}{@{}l@{\hspace{3pt}}ccccc@{}}
    & {\scriptsize Ground Truth} & {\scriptsize D3G}
      & {\scriptsize STG} & {\scriptsize 4DSG}
      & {\scriptsize Ours} \\
    \rotatebox{90}{\scriptsize \hspace{4pt}$t=2$}
      & \includegraphics[width=\shw]{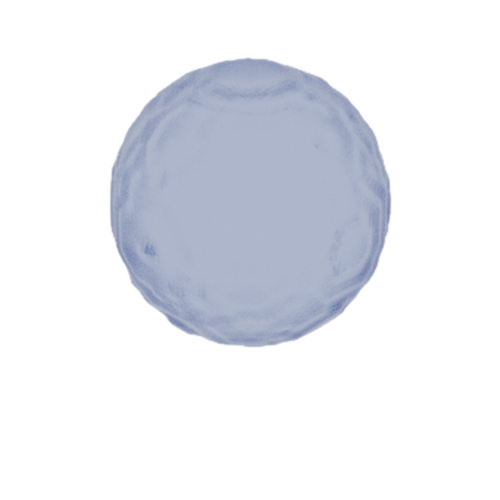}
      & \includegraphics[width=\shw]{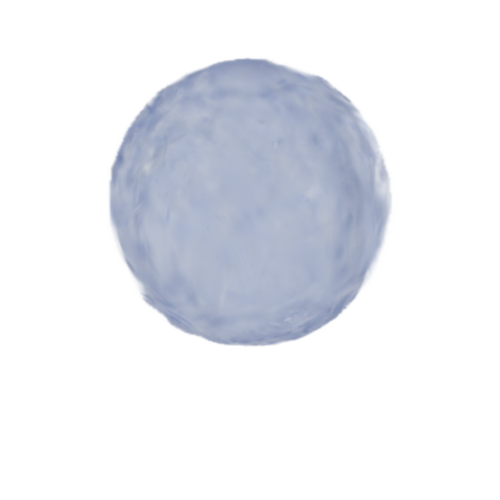}
      & \includegraphics[width=\shw]{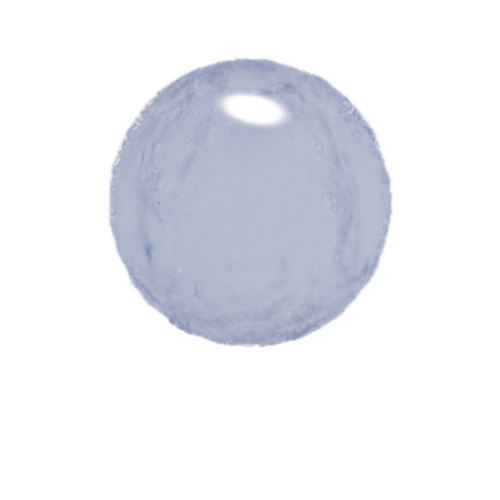}
      & \includegraphics[width=\shw]{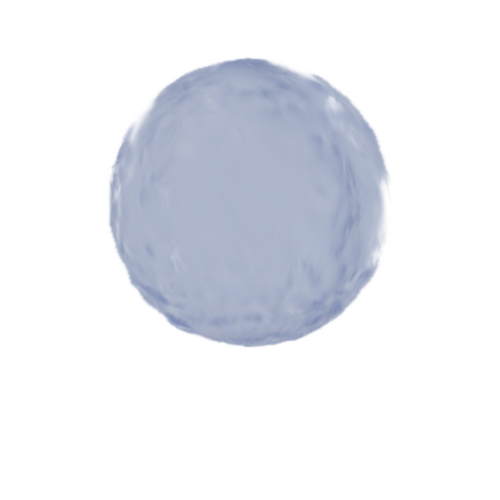}
      & \includegraphics[width=\shw]{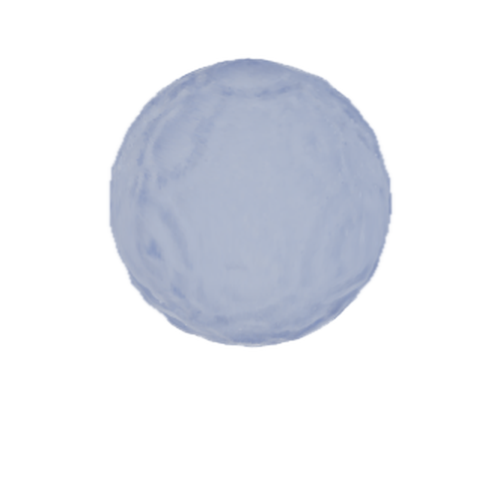} \\
    \rotatebox{90}{\scriptsize \hspace{4pt}$t=8$}
      & \includegraphics[width=\shw]{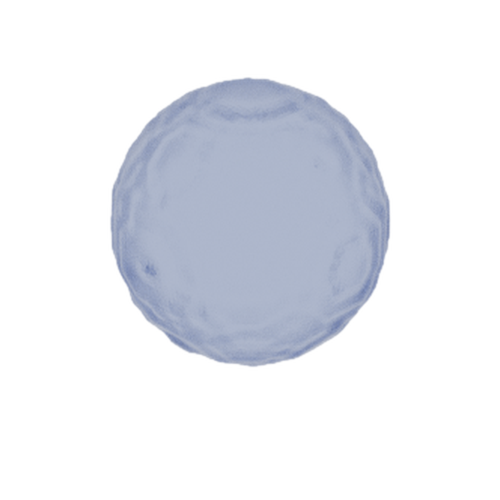}
      & \includegraphics[width=\shw]{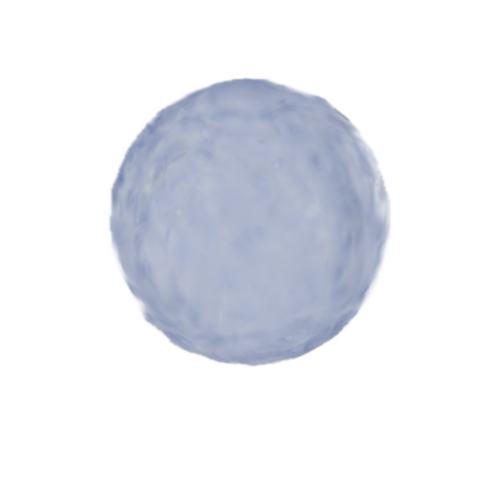}
      & \includegraphics[width=\shw]{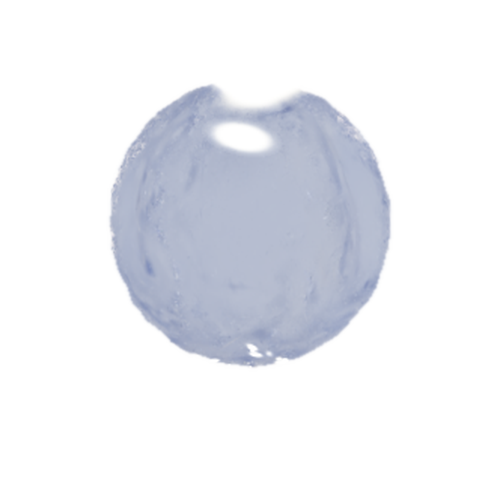}
      & \includegraphics[width=\shw]{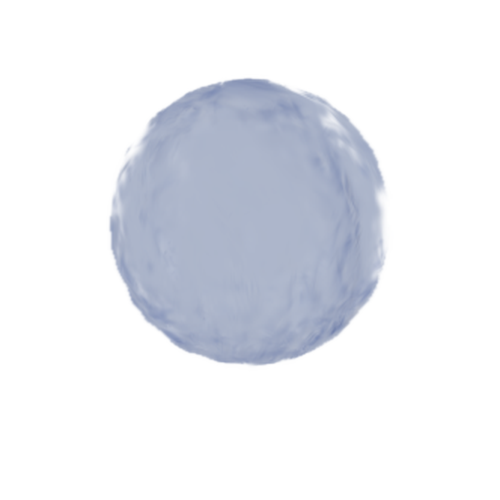}
      & \includegraphics[width=\shw]{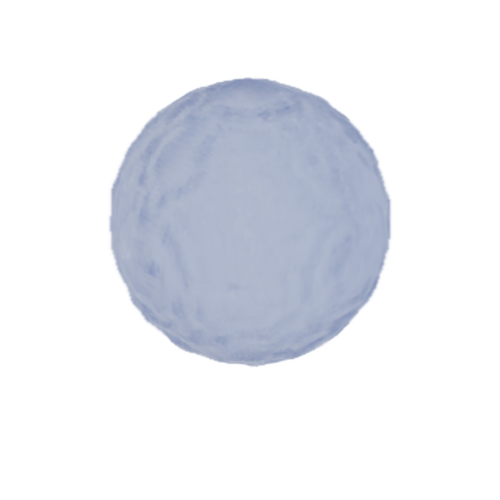} \\
    \rotatebox{90}{\scriptsize \hspace{2pt}$t=14$}
      & \includegraphics[width=\shw]{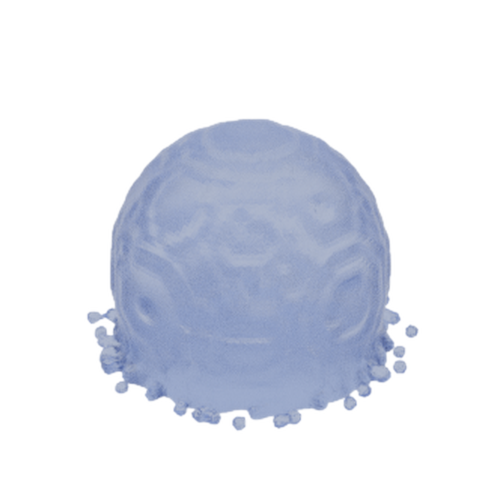}
      & \includegraphics[width=\shw]{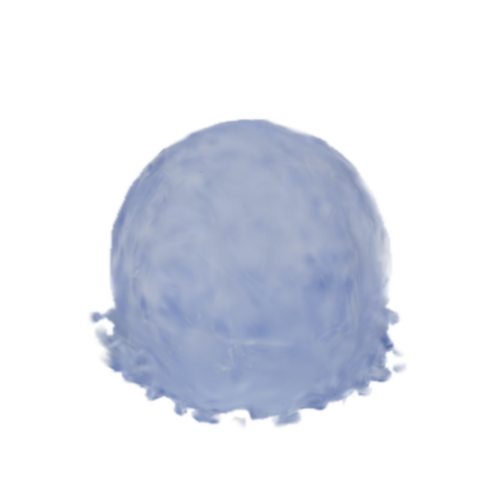}
      & \includegraphics[width=\shw]{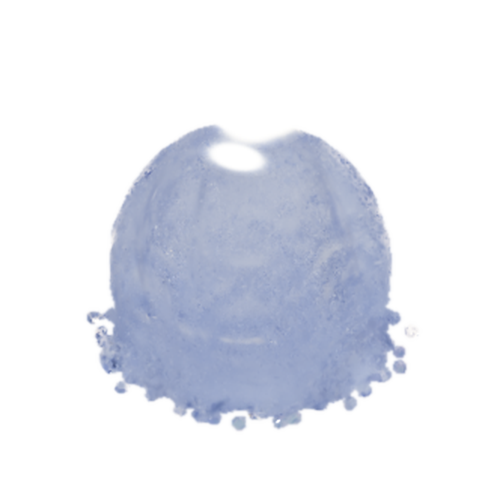}
      & \includegraphics[width=\shw]{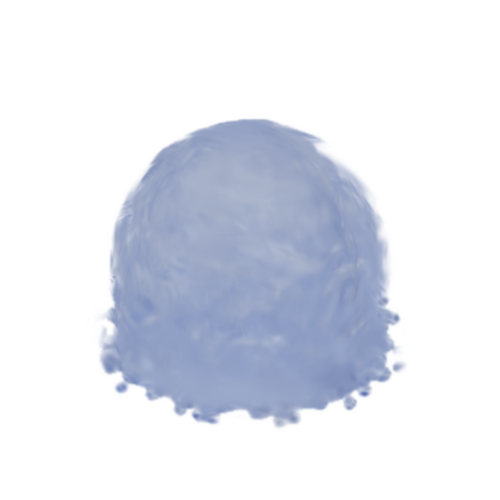}
      & \includegraphics[width=\shw]{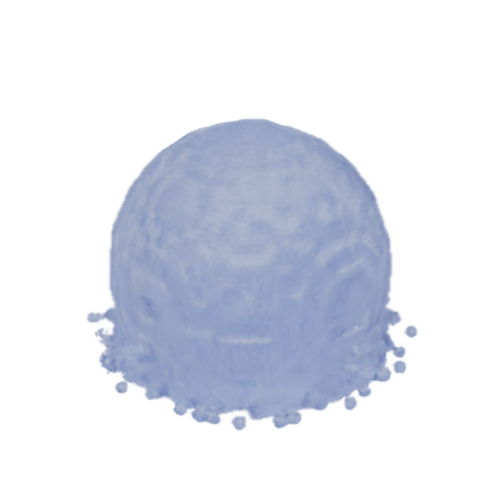} \\
    \rotatebox{90}{\scriptsize \hspace{2pt}$t=19$}
      & \includegraphics[width=\shw]{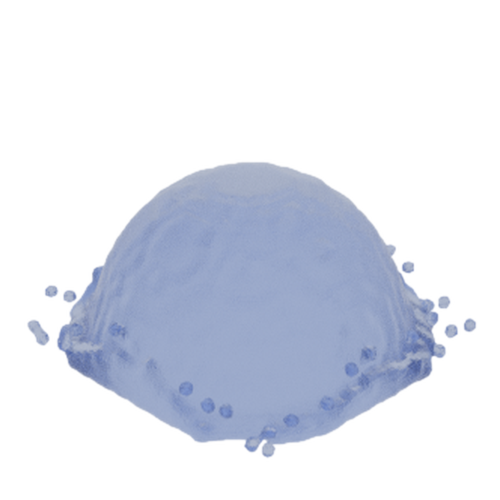}
      & \includegraphics[width=\shw]{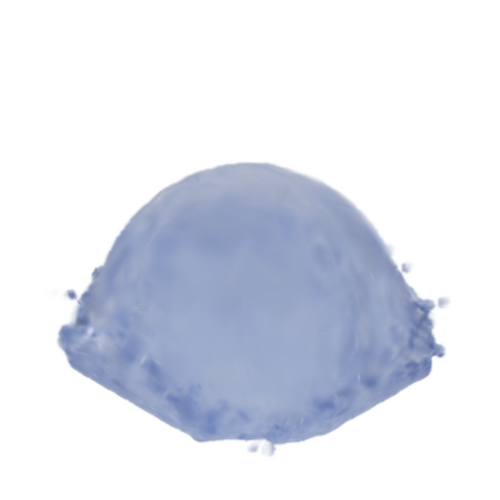}
      & \includegraphics[width=\shw]{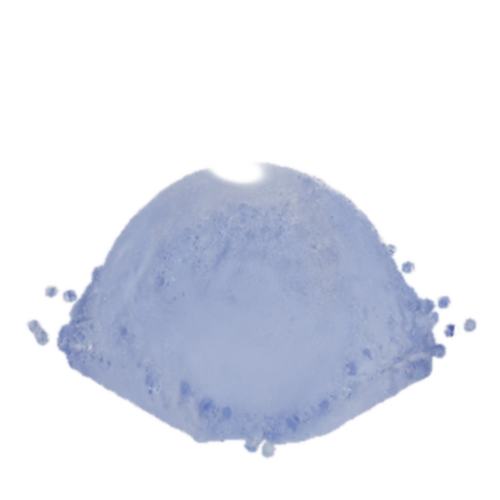}
      & \includegraphics[width=\shw]{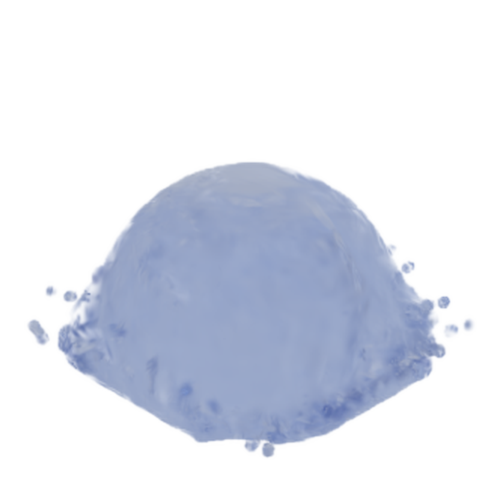}
      & \includegraphics[width=\shw]{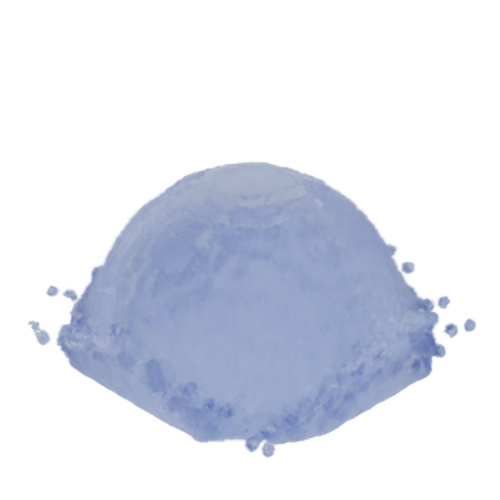} \\
  \end{tabular}
  \caption{\textbf{Novel view synthesis on NeuroFluid
\emph{WaterSphere}.} The baselines remain plausible while the surface is smooth but degrade as it deforms, losing rim structure and detached droplets at impact; SpacetimeGaussians additionally bakes a static specular highlight into the moving surface.}
  \label{fig:watersphere_showcase}
\end{figure}

\vspace{1mm}\noindent\textbf{Novel view synthesis.}
\cref{fig:crop_showcase} compares novel-view synthesis on our real
captures. On training views all methods reproduce the scene
reasonably, but the baselines already blur or drop the thin pouring
stream, the structure with the least photometric support. The gap
widens on test views, the baselines place liquid Gaussians away from the observed surface, which fills the scene with haze and occludes the container, since a single dynamic representation has nothing holding the liquid to its own region. \methodname{} keeps the stream continuous and the liquid confined to the observed surface.
\cref{tab:nvs_fg} presents the quantitative comparison, and shows the trade-off anticipated in \cref{sec:intro}: 4D-Scaffold-GS attains the best baseline PSNR but the largest density deviation, while Deformable-3DGS is the most physically consistent of the three and the weakest photometrically. \methodname{} bypasses this trade-off. By advecting carriers along a velocity field fitted to the observed interface while preserving coverage across observations, motion is resolved prior to appearance optimization, leaving the decoder to model only residuals. Consequently, \methodname{} attains the best novel-view quality together with the lowest density deviation and energy drift, and reduces temporal jitter by a third relative to the strongest baseline. The sparse carrier representation is also the cheapest to fit, with the lowest training time and peak memory.

The synthetic \emph{WaterSphere} scene~\cite{guan2022neurofluid}
isolates the effect without the appearance ambiguity of real water
(\cref{fig:watersphere_showcase}). On synthetic renders the silhouette is trivially recoverable from the images and carries no
information beyond the photometric signal, so no method holds privileged input. The baselines remain plausible at early timesteps but degrade as the surface deforms, losing the rim structure and detached droplets at impact, and STG bakes a static specular
highlight into the moving surface. \methodname{} tracks the
deformation throughout and attains the best PSNR, LPIPS, and temporal
jitter (\cref{tab:watersphere_fg}).

\begin{table}[t]
\centering
\scriptsize
\caption{\textbf{Foreground novel-view synthesis on NeuroFluid
\emph{WaterSphere}}~\cite{guan2022neurofluid}, over all 61 held-out
frames.}
\label{tab:watersphere_fg}
\begin{tabular}{l cccc}
\toprule
Method & PSNR$\uparrow$ & SSIM$\uparrow$ & LPIPS$\downarrow$
& $|\Delta\mathrm{Jitter}|\downarrow$ \\
\midrule
D3G~\cite{yang2023deformable3dgs} & 25.87 & \textbf{0.8763} & 0.2670 & 0.0239 \\
STG~\cite{Li_STG_2024_CVPR}       & 22.05 & 0.8178 & 0.3523 & 0.0307 \\
4DSG~\cite{cho20264d}             & 26.10 & 0.8730 & 0.2538 & 0.0213 \\
\midrule
\textbf{Ours} & \textbf{27.94} & 0.8761 & \textbf{0.2329} & \textbf{0.0099} \\
\bottomrule
\end{tabular}
\vspace{-2pt}
\end{table}

\begin{figure}[t]
  \centering
  \setlength{\tabcolsep}{0pt}
  \newcommand{\shw}{0.16\linewidth}
  \begin{tabular}{@{}cccccc@{}}
    \scriptsize Reference & \scriptsize Capture $t$ & \scriptsize 4DSG
      & \scriptsize Ours & \scriptsize Ground truth & \scriptsize Capture $t\!+\!1$ \\
      \includegraphics[width=\shw]{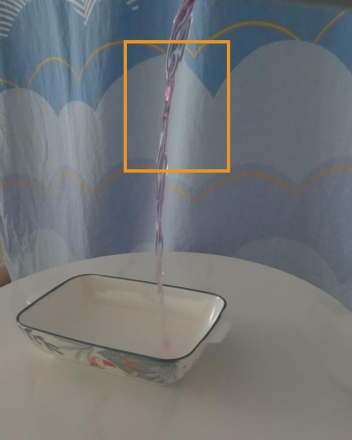}
      & \includegraphics[width=\shw]{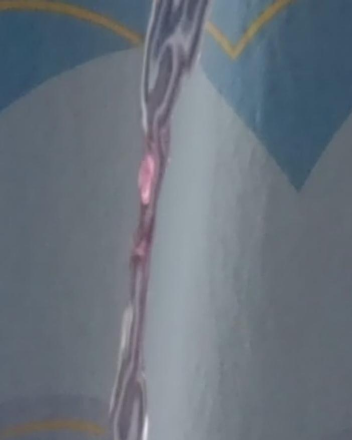}
      & \includegraphics[width=\shw]{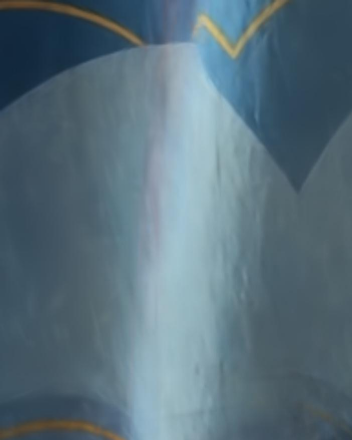}
      & \includegraphics[width=\shw]{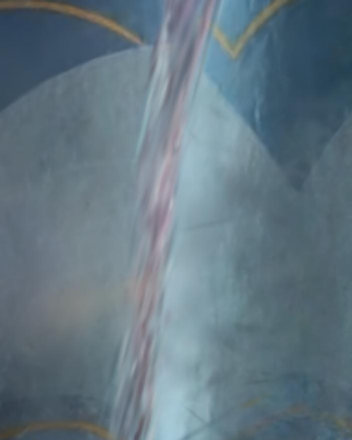}
      & \includegraphics[width=\shw]{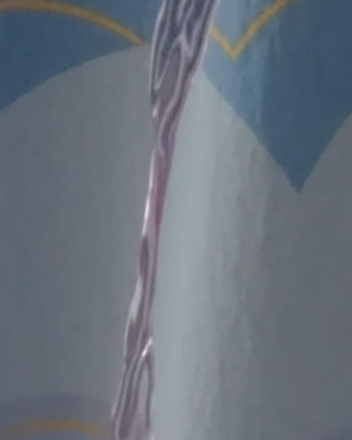}
      & \includegraphics[width=\shw]{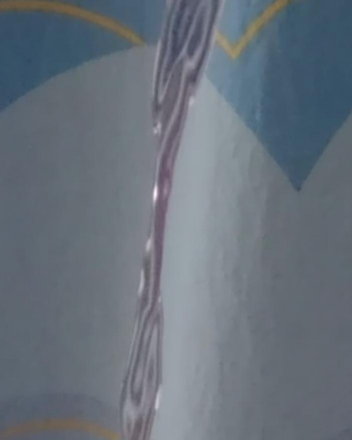} \\
      \includegraphics[width=\shw]{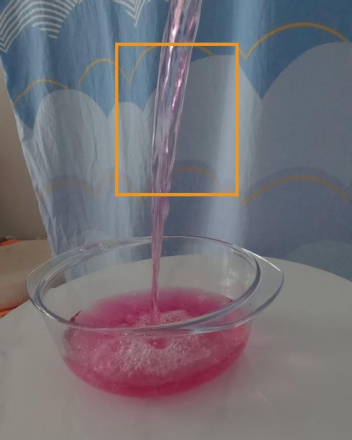}
      & \includegraphics[width=\shw]{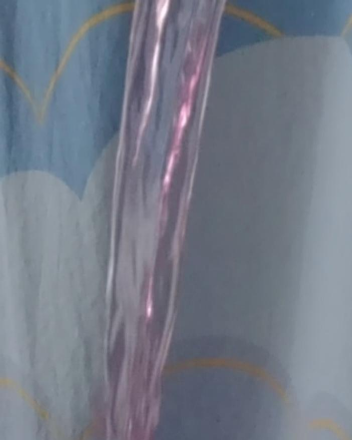}
      & \includegraphics[width=\shw]{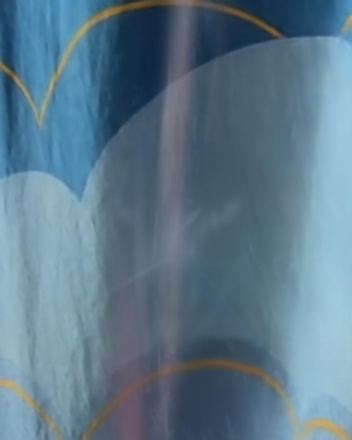}
      & \includegraphics[width=\shw]{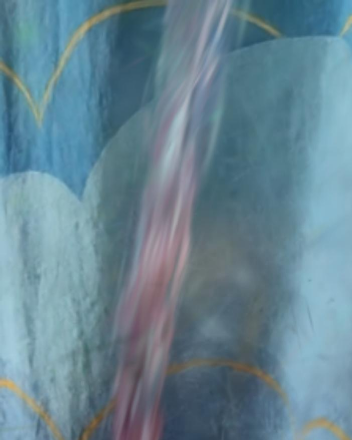}
      & \includegraphics[width=\shw]{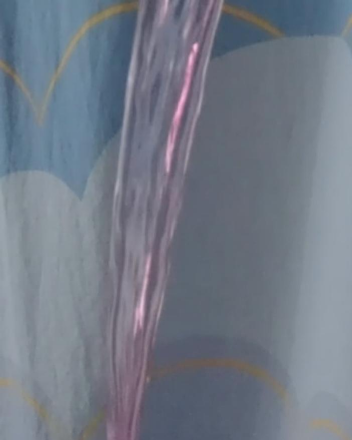}
      & \includegraphics[width=\shw]{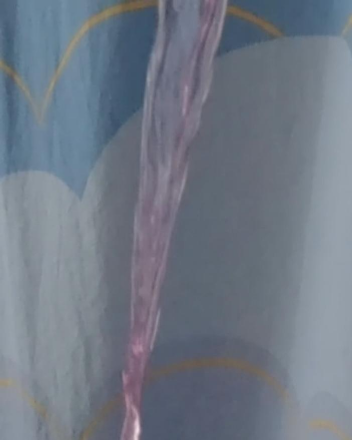} \\
  \end{tabular}
\caption{\textbf{Temporal interpolation.} Every method is fitted to the even captured frames only, so the frame in the middle columns was
withheld during training and serves as ground truth. The outer columns are the two captures that bracket it, and the amber box in
the reference marks the crop. 4D-Scaffold-GS attenuates the falling stream into a faint band, while ours preserves its colour, thickness, and position.}
  \label{fig:interp}
\end{figure}

\vspace{1mm}\noindent\textbf{Temporal interpolation.}
Fitting a sequence well does not require having recovered its motion: the observed frames can be explained in many ways. Rendering at an
instant that was never trained on separates the two, so we fit our method and strongest baseline \cite{cho20264d} to the even frames only and compare against the withheld ones. A
deformation field can be queried there too, but it interpolates appearance, whereas advecting carriers transports material.
\cref{fig:interp} shows the qualitative comparison: 4DSG attenuates the falling stream into a faint band, while ours keeps its
colour and thickness and lands where the withheld capture places it.

\begin{figure}[t]
  \centering
  \setlength{\tabcolsep}{0pt}
  \newcommand{\stw}{0.25\linewidth}
  \begin{tabular}{@{}cccc@{}}
    \scriptsize Reconstructed & \scriptsize Clear water & \scriptsize Brown dye & \scriptsize Blue dye \\
    \includegraphics[width=\stw]{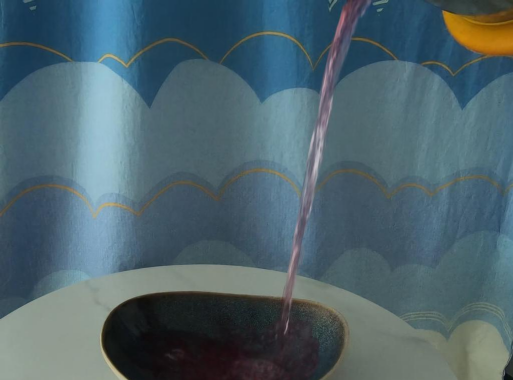}
    & \includegraphics[width=\stw]{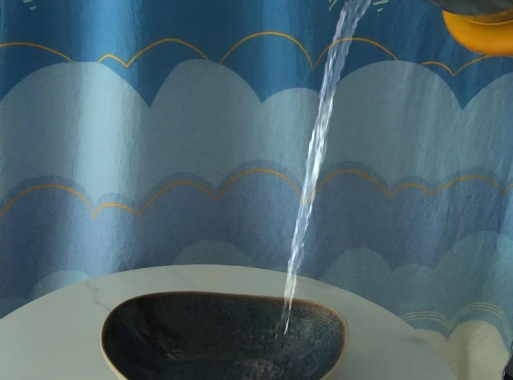}
    & \includegraphics[width=\stw]{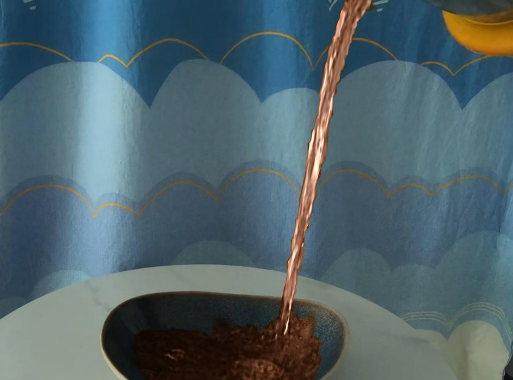}
    & \includegraphics[width=\stw]{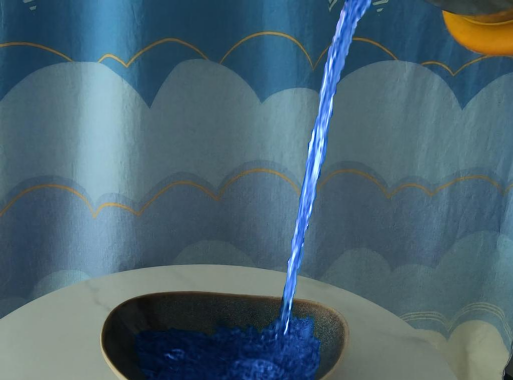} \\
    \includegraphics[width=\stw]{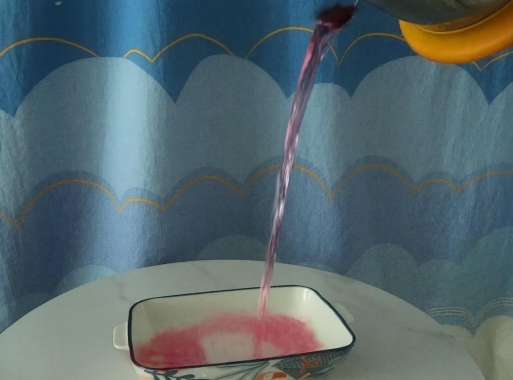}
    & \includegraphics[width=\stw]{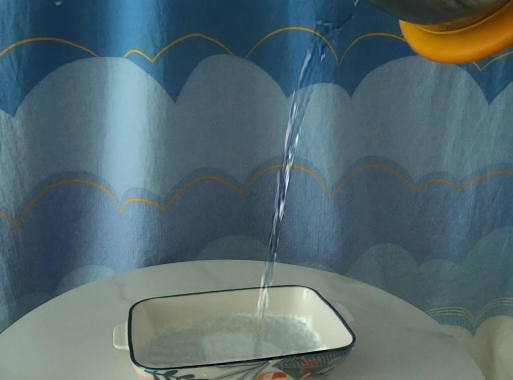}
    & \includegraphics[width=\stw]{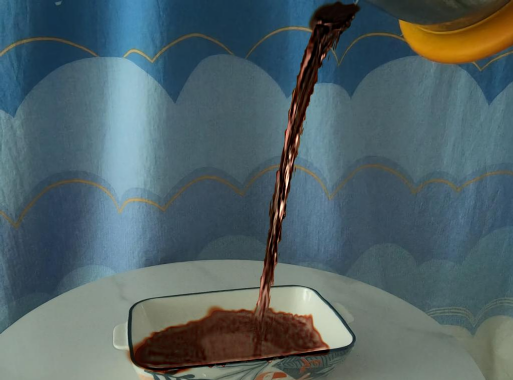}
    & \includegraphics[width=\stw]{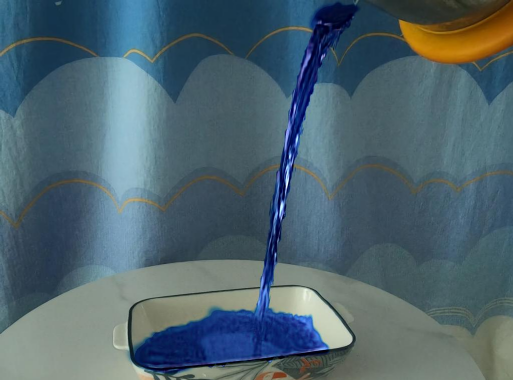} \\
  \end{tabular}
    \caption{\textbf{Style transfer.} New optical properties are assigned to the reconstructed liquid and light transport is re-solved against the scene behind it, with no mesh extraction and no re-optimization.
Refraction of the background and the ripples of the recording are preserved.}
  \label{fig:style}
\end{figure}

\noindent\textbf{Style transfer.}
Because our carriers are placed by the observed geometry rather than by appearance, the liquid's optical properties are not entangled with its shape: the SDF and the carrier positions stay fixed while the
material is replaced. Re-rendering then needs only what the reconstruction already provides, the optical path length along a ray
and the orientation of the interface, so light transport can be re-solved against the frozen background without extracting a mesh or
re-optimizing anything~\cite{max1995optical, schlick1994fresnel, zhang2025stylizedgs}.
Removing the dye leaves clear water through which the vessel stays visible, while absorbing dielectrics deepen in colour with
accumulated volume, and the stream keeps the width, ripples, and highlights of the recording (\cref{fig:style}).

\begin{figure}[t]
  \centering
  \includegraphics[width=0.98\linewidth]{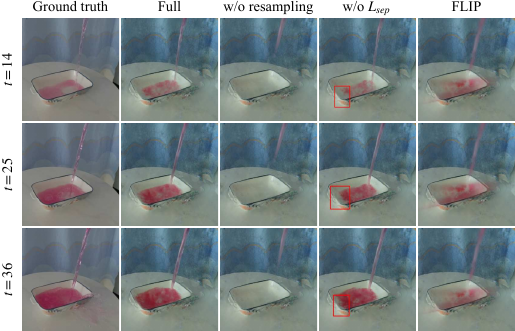}
  \caption{\textbf{Qualitative ablation on test views.} Without resampling the stream degrades into a diffuse veil and the accumulated liquid disappears by the end of the sequence; without $\mathcal{L}_{\mathrm{sep}}$ the thin early stream is washed out; and open-loop FLIP transports the liquid toward the bottom of the bowl but drifts from the observed surface. The full model preserves both the stream and accumulated liquid.}
  \label{fig:ablation_showcase_test}
\end{figure}

 \subsection{Ablation Study}
\label{sec:ablation}
We ablate on 10 scenes from our real captures. Since the liquid is largely
transparent, pixel error within the foreground mask remains dominated
by the static scene visible through it, and PSNR spans only 2.84\,dB
between our full model and a rendering with no liquid at all. We therefore also report mIoU@0.5 between the annotated liquid mask and
the thresholded rendered opacity, which measures whether the reconstruction occupies the observed liquid region. We compare the full model against four variants: \emph{w/o resampling}, where carriers are never reseeded
after initialization; \emph{w/o $\mathcal{L}_{\mathrm{sep}}$},
without the separation term between the $K$ children of a carrier;
\emph{Per-frame}, where carriers are reseeded from the SDF at every
frame without transport; and \emph{FLIP}, where they are advected by
an open-loop solver instead of the fitted velocity field. 
Further ablations are
in the supplementary material.

\begin{table}[t]
\centering
\caption{\textbf{Ablation on our real captures.} Foreground metrics on
held-out views, averaged over ten scenes. mIoU@0.5 is the IoU between the annotated liquid mask and the foreground mask obtained by
thresholding rendered foreground-Gaussian opacity at 0.5; \# Carr. is the mean
number of active carriers per frame.}
\label{tab:ablation}
\scriptsize
\setlength{\tabcolsep}{4pt}
\begin{tabular}{l cccc c}
\toprule
Variant & mIoU@0.5$\uparrow$  & PSNR$\uparrow$ & LPIPS$\downarrow$
& $\sigma_D\downarrow$ 
& \# Carr. \\
\midrule
w/o resampling &0.3116 & 19.05 & 0.093 & 384.7 & 0.8k \\
Per-frame   &0.7598   & 20.06 & 0.077 & 243.3 & 56.4k\\
FLIP        &0.2999	   & 18.26 & 0.091 & 784.3 & 0.4k \\
\midrule
\textbf{Full} & \textbf{0.7696} & \textbf{20.31} & \textbf{0.076}
& \textbf{234.2} & 26.7k  \\
\bottomrule
\end{tabular}
\end{table}

\vspace{1mm}\noindent\textbf{Effect of resampling.}
An opaque liquid exposes only its free surface, so the mask-fused SDF is the primary evidence of where material exists, and resampling is
what converts this evidence into foreground Gaussians. Without it, carrier deaths are not replenished and the active population falls
from 26.7k to 0.8k, and every metric degrades. The stream thins into a diffuse veil and the accumulated water body has largely disappeared by the end of the sequence (\cref{fig:ablation_showcase_test}). Coverage of the observed liquid is therefore not something transport can maintain on its own; it has to be restored from each new observation.

\vspace{1mm}\noindent\textbf{Role of Transport.}
The \emph{Per-frame} variant reaches comparable image quality, which
is expected, since the per-frame SDF already pins the geometry. It
does so with 56.4k carriers against our 26.7k and a less uniform
distribution, as each frame rediscovers the liquid rather than inheriting it. Advecting carriers along the fitted field keeps the coverage compact, and since their
motion is defined between observations, the representation can also
be decoded at sub-frame times, which per-frame reseeding cannot do
(\cref{fig:interp}).

\vspace{1mm}\noindent\textbf{Child separation.}
\emph{w/o $\mathcal{L}_{\mathrm{sep}}$} leaves the carrier pool and its density uniformity unchanged, since the term acts only within a
carrier to prevent its child Gaussians from collapsing. Its effect is visible in the accumulated liquid in \cref{fig:ablation_showcase_test}, which
renders thinner and less saturated when the $K$ children overlap instead of covering distinct volume.

\vspace{1mm}\noindent\textbf{Physics-constrained optimization vs. physical simulation.}
The \emph{FLIP} variant produces large degradation, in both appearance and density uniformity. The solver is not at fault: it runs without the inflow rate, total volume, and boundary conditions that images do not reveal, so its motion is internally consistent but unrelated to the observed liquid, and the photometric objective cannot recover from the mismatch. \cref{fig:ablation_showcase_test}
shows the mechanism: Gaussians are transported toward the bottom of the bowl but drift from the observed surface once nothing corrects them. A simulator imposes physical structure everywhere, including where evidence is insufficient, and our loop constrains only observation-supported area.

\section{Conclusion}
We presented \methodname{}, a benchmark and a method for
reconstructing splashing liquids from real-world multi-view video.
Reconstruction research on fluids has so far relied on smoke or
simulation. Our benchmark provides synchronized, calibrated captures
of real liquids in violent motion, and our method shows the setting
is tractable when physical structure is imposed only where the
observations support it. \methodname{} reconstructs real splashes
more accurately and with more plausible motion than dynamic Gaussian
splatting baselines, at a lower training cost, and the same
representation supports temporal interpolation and style transfer. 

\vspace{1mm}\noindent\textbf{Limitations.}
\methodname{} reconstructs the liquid from silhouettes, which
constrain coarse surface geometry and interface motion but not the specular glints and
refraction that give water its character. Transient structures such
as impact bubbles and foam are also missed, being too small and
short-lived to be carved from multi-view masks. Our velocity field is
likewise kinematic: consistent with the observed interface, but not
with a momentum balance. Recovering interpretable dynamics from real
captures of this kind remains open, and is the direction this
benchmark is built to support.

{
    \small
    \bibliographystyle{ieeenat_fullname}
    \bibliography{main}
}

\clearpage
\setcounter{page}{1}
\renewcommand{\thepage}{S\arabic{page}}
\maketitlesupplementary

\setcounter{table}{0}
\renewcommand{\thetable}{S\arabic{table}}
\setcounter{figure}{0}
\renewcommand{\thefigure}{S\arabic{figure}}
\setcounter{equation}{0}
\renewcommand{\theequation}{S\arabic{equation}}
\setcounter{section}{0}
\renewcommand{\thesection}{S\arabic{section}}

In this supplementary material, we provide additional details on the benchmark (Sec.~\ref{sec:supp_benchmark}), more implementation and evaluation details (Sec.~\ref{sec:supp_method}), together with additional comparison results (Sec.~\ref{sec:supp_experiments}), and extended ablation studies (Sec.~\ref{sec:supp_ablations})
that complement the main paper.
Please note that our supplementary video results
(\texttt{splashsplat\_video.mp4}) are provided as a separate file
accompanying this submission.

\section{Benchmark Details}
\label{sec:supp_benchmark}
 
\subsection{Capture Setup}
Seven GoPro cameras record each of the 20 scenes at $3840\times2160$
and $60$\,fps, arranged along an arc spanning $77^\circ$ on average
around the scene center; each held-out camera is $8^\circ$--$15^\circ$
from its nearest training view. Since the cameras free-run, we
synchronize each capture session offline using audio. Specifically, we
estimate each camera's temporal offset to a reference by
cross-correlating loudness envelopes and apply the resulting whole-frame
shift to all sequences from that session. The half-frame figure reported
in the main paper characterizes the quantization introduced by this
whole-frame alignment: at $60$\,fps, rounding contributes at most
$8.33$\,ms, corresponding to $4.8$\,ms RMS under a uniform rounding
error.

Beyond this quantization term, residual misalignment can arise from
offset-estimation uncertainty and clock-rate drift, so the measured
sequence-level residual can exceed $8.33$\,ms. Two cameras exhibit
measured rate offsets of $-49\,\mu\mathrm{s/s}$ and
$-81\,\mu\mathrm{s/s}$, which cannot be corrected by a single
session-level shift. Over the longest $640$\,s session, the larger
drift accumulates to $52$\,ms, or approximately three source frames.
Consequently, the ten scenes extracted from this session have a higher
mean residual synchronization error ($7.5$\,ms) than the ten scenes
from shorter sessions ($4.1$\,ms).

To quantify the residual synchronization of each released sequence, we
re-estimate the camera offsets within its own temporal window using
GCC-PHAT~\cite{1162830}. We report the RMS of the seven offsets after
centering them by their mean as \emph{Sync RMS} in
Tab.~\ref{tab:supp_scenes}. The median Sync RMS across the dataset is
$5.2$\,ms; 17 of the 20 scenes are below $8.33$\,ms, while the three
higher-error sequences are marked with $\ddagger$.

\subsection{Scene Overview}
Tab.~\ref{tab:supp_scenes} lists all 20 scenes. Every sequence is 7 cameras $\times$ 50 timesteps. \emph{Interval} is the wall-clock span
the 50 timesteps cover and \emph{Eff.\ $\Delta t$} their mean spacing, so the 50 timesteps subsample the 60\,fps capture to an
effective 9--16\,fps, chosen to span the full deformation rather than
a short burst of it; the sampling step is not an integer number of source frames, so consecutive gaps alternate about the mean by up to half a source frame. \emph{Sync RMS} is the residual synchronization
error defined in Sec.~\ref{sec:supp_benchmark}. In three sequences it leaves one camera a full source
frame or more from the reference ($\ddagger$) --- the clock drift above at its worst;
these are released and evaluated exactly as trained on, and flagged so the effect is
visible. Fig.~\ref{fig:supp_gallery} shows one representative frame per scene.
 
\begin{table*}[t]
  \centering
  \caption{\textbf{Benchmark scenes.} The 20 sequences, with the wall-clock interval their 50 timesteps span, the mean spacing between them, and the residual synchronization error across the seven cameras. $\dagger$: fewer than 7 cameras could be measured. $\ddagger$: one camera is a full source frame or more from the reference.}
  \label{tab:supp_scenes}
  \small
  \begin{tabular}{l l r r r@{\hspace{2em}}l l r r r}
    \toprule
    Scene & Container & Interval & Eff.\ $\Delta t$ & Sync RMS & Scene & Container & Interval & Eff.\ $\Delta t$ & Sync RMS \\
     &  & (s) & (ms) & (ms) &  &  & (s) & (ms) & (ms) \\
    \midrule
    bowl\_001 & Bowl & 5.34 & 109.0 & 5.2 & bowl\_011 & Bowl & 3.34 & 68.1 & 5.4$^{\dagger}$ \\
    bowl\_002 & Bowl & 3.84 & 78.3 & 4.1$^{\dagger}$ & bowl\_012 & Bowl & 4.34 & 88.5 & 7.0$^{\dagger}$ \\
    bowl\_003 & Bowl & 3.92 & 80.0 & 2.4$^{\dagger}$ & bowl\_013 & Bowl & 3.85 & 78.6 & 7.7 \\
    bowl\_004 & Bowl & 4.57 & 93.3 & 2.1 & bowl\_014 & Bowl & 5.17 & 105.5 & 9.3$^{\ddagger}$ \\
    bowl\_005 & Bowl & 5.22 & 106.6 & 5.3 & bowl\_015 & Bowl & 3.97 & 81.0 & 11.7$^{\dagger}$$^{\ddagger}$ \\
    bowl\_006 & Bowl & 4.84 & 98.7 & 5.2 & bowl\_016 & Bowl & 3.00 & 61.3 & 3.5 \\
    bowl\_007 & Bowl & 3.67 & 74.9 & 8.1$^{\dagger}$ & bowl\_017 & Bowl & 3.54 & 72.2 & 3.7 \\
    bowl\_008 & Bowl & 3.22 & 65.7 & 9.4$^{\dagger}$$^{\ddagger}$ & tank\_001 & Tank & 4.19 & 85.5 & 4.5$^{\dagger}$ \\
    bowl\_009 & Bowl & 4.00 & 81.7 & 6.3 & tank\_002 & Tank & 4.49 & 91.6 & 3.5$^{\dagger}$ \\
    bowl\_010 & Bowl & 4.70 & 96.0 & 4.6 & tank\_003 & Tank & 3.67 & 74.9 & 6.9$^{\dagger}$ \\
    \bottomrule
  \end{tabular}
\end{table*} 
 
\subsection{Annotation Protocol}
We compare the uncorrected SAM3 output and against the refined, annotated masks. The mean IoU is $0.772$ and the median $0.967$: most masks are accepted almost unchanged and a
minority require most of the refinement effort. Correction is almost entirely additive, $20.1\%$ of the union added against $2.7\%$ removed, and $8.6\%$ of the raw masks are empty, i.e.\ SAM3 returns nothing at all. Averaged
over a frame the editing touches $1.00\%$ of the pixels and raises mask coverage from $1.89\%$ to $2.70\%$. The effort is far from uniform: per-scene mean IoU ranges from $0.36$ to $0.96$, lowest for
dyed liquid seen through glass, where SAM3 most often does not segment the  liquid.

\begin{figure}[t]
  \centering
  \includegraphics[width=\linewidth]{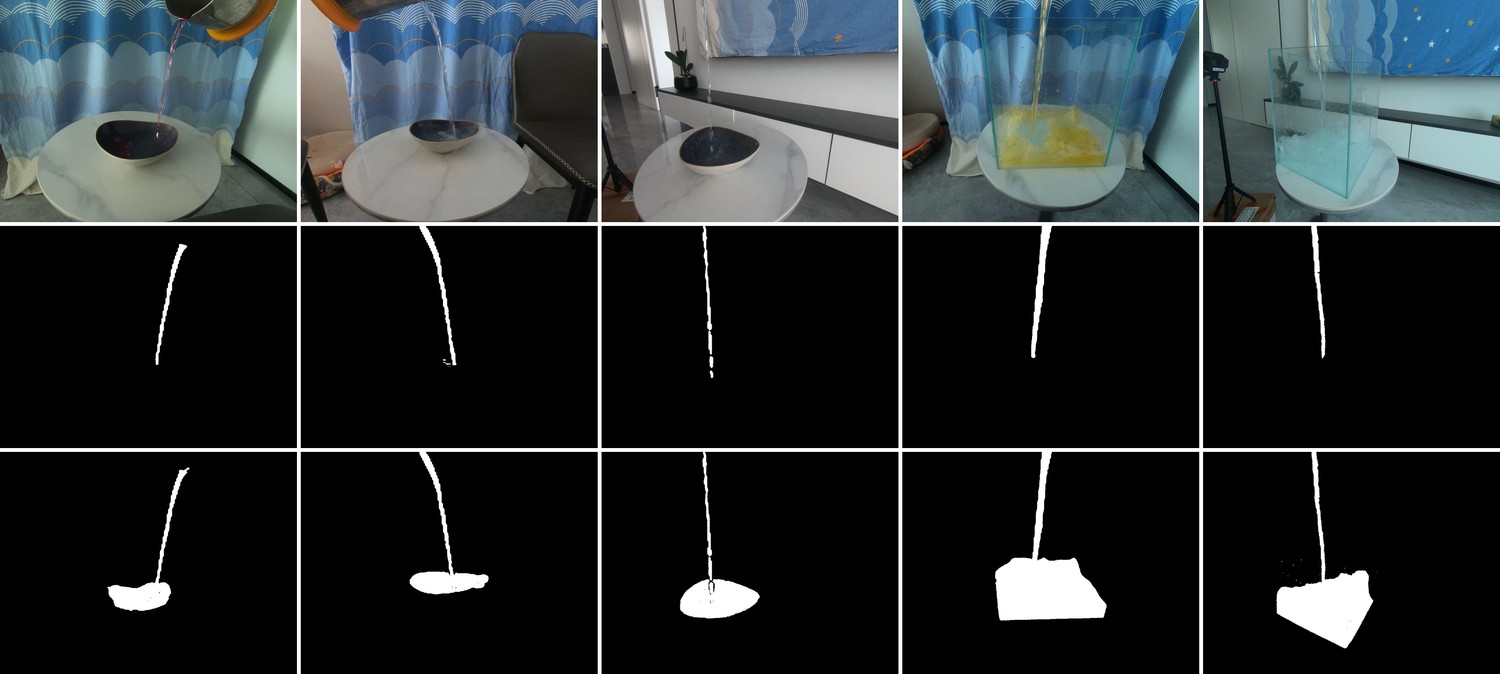}
  \caption{\textbf{Raw SAM3 versus refined masks.} Five frames, ordered by increasing
  disagreement. Top: frame. Middle: SAM3 at confidence $0.5$. Bottom: the released mask.
  Columns 1--3 miss the pooled liquid in an opaque bowl, and columns 4--5 recover the stream but not the water standing in the tank.}
  \label{fig:supp_masks}
\end{figure}

\begin{figure*}[t]
  \centering
  \includegraphics[width=\linewidth]{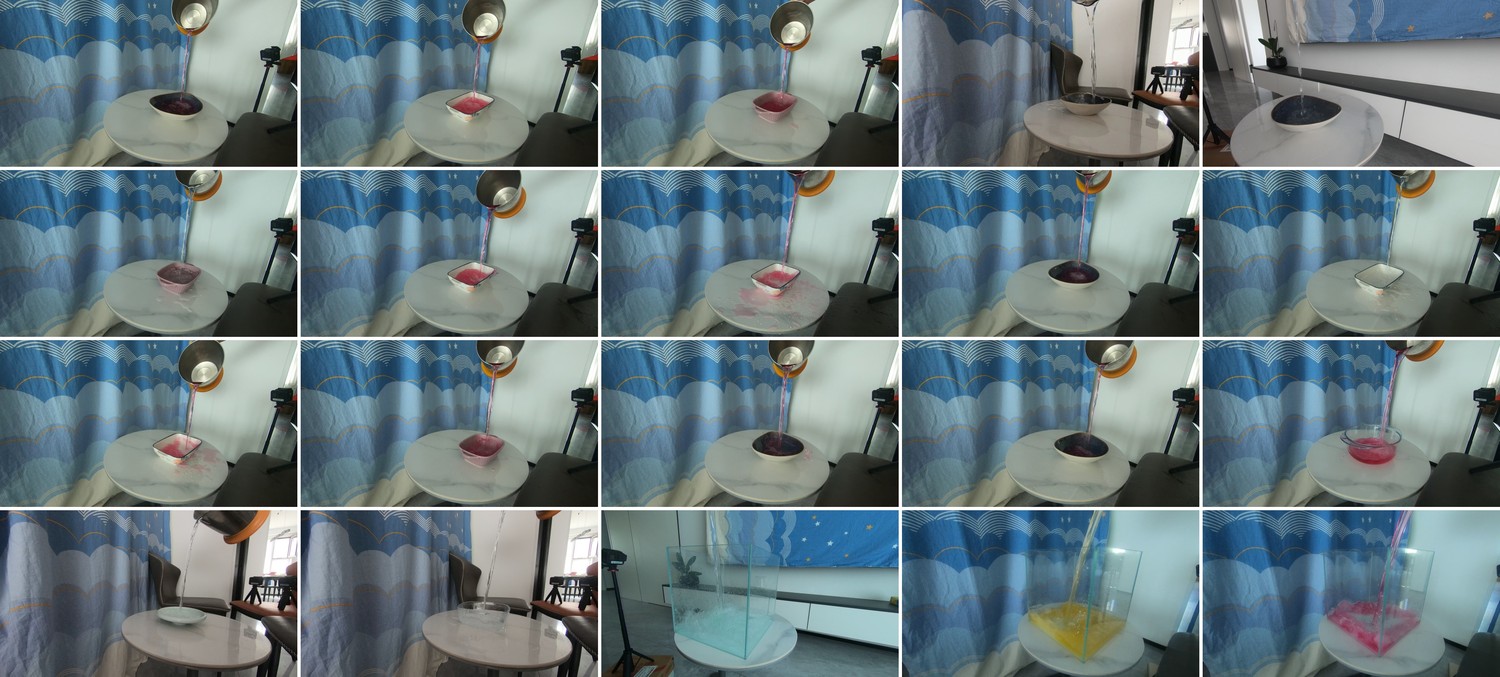}
  \caption{\textbf{Scene gallery.} One representative frame from each of the 20 benchmark scenes.}
  \label{fig:supp_gallery}
\end{figure*}

Fig.~\ref{fig:supp_masks} shows what the correction consists of. Against an opaque container SAM3 recovers the falling stream but stops at the liquid surface; behind a glass wall it recovers the stream but almost none of the water standing below it. 
 
To extend the dataset modalities and its possible applications, we scan the three primary containers used in our captures with AR Code~\cite{arcode2024}. The reconstructed meshes are imported into Blender, where we manually remove scanning artifacts, fill missing surface regions, and refine noisy boundaries to obtain clean, watertight geometry. We then align the processed meshes with the capture setup and verify their geometry by projecting them into the calibrated multi-view images.

\subsection{Evaluation Protocol}

Every scene uses the same seven-camera rig, with cameras 1--5 for training and 6--7 held out; both test views are bracketed by training cameras. Each camera is cropped to a window fixed in size and position:
the union of its mask bounding boxes over all frames and both splits, padded by $96$\,px and rounded up to a multiple of eight. Only the principal point is shifted, and all methods are trained and scored on
these crops. 

All methods are exported as per-frame positions and activity flags and scored by one routine. Our carriers represent fluid by construction; for baselines that mix background and fluid, we keep only primitives projecting inside the ground-truth silhouette in all
training views, a conservative criterion that excludes background Gaussians at the cost of discarding fluid primitives visible in only part of the rig. $\sigma_D$ is the per-frame standard deviation of an
SPH density estimate, using a Poly6 kernel of radius $0.3$ over the
$64$ nearest neighbours with unit mass and the self term excluded,
averaged over frames; primitive counts are capped by subsampling so
that the statistic is comparable across methods whose counts differ
by orders of magnitude. $\Delta E$ averages the mechanical energy of
carriers alive in consecutive frames, from finite-difference
velocities and a gravitational potential. Both statistics depend on
particle count, and since our resampling maintains near-uniform
coverage, $\sigma_D$ partially reflects our design; we therefore also
report the normalized $\sigma_D/\mu_D$.

\section{Method and Training Details}
\label{sec:supp_method}
 
\subsection{Loss Definitions and Weights}

We use the notation of Sec.~4: $\hat I_{t,v}$ is the composite of
$\mathcal{G}^{\mathrm{fg}}_t$ over the frozen $\mathcal{G}^{\mathrm{bg}}$ (Eq.~1),
$\hat\alpha_{t,v}$ the accumulated opacity obtained by rasterizing
$\mathcal{G}^{\mathrm{fg}}_t$ alone under $\Pi_v$, $M_{t,v}$ the binary liquid mask,
$\Omega$ the pixel domain, and $\mathcal{A}_t$ the set of active carriers. All image terms
are computed per sampled pair $(t,v)$.

\paragraph{Photometric terms.}
The foreground-normalized objective is
\begin{equation}
    \mathcal{L}_{\mathrm{fg}}
    =
    \frac{\sum_{p\in\Omega} M_{t,v}(p)\,\big\|\hat I_{t,v}(p)-I_{t,v}(p)\big\|_1}
         {3\sum_{p\in\Omega} M_{t,v}(p)+\epsilon},
    \label{eq:foreground_loss}
\end{equation}
with $\epsilon=10^{-6}$. Dividing by the masked area rather than the image makes a thin jet
and a wide splash contribute equally. The gradient term applies the same normalization to
finite differences $\nabla_{pq}I=I(q)-I(p)$ over adjacent pixel pairs $(p,q)$ along the
image axes,
\begin{equation}
\begin{aligned}
    \mathcal{L}_{\mathrm{grad}}
    &=
    \frac{\sum_{(p,q)} w_{pq}\,\big\|\nabla_{pq}\hat I_{t,v}-\nabla_{pq} I_{t,v}\big\|_1}
         {3\sum_{(p,q)} w_{pq}+\epsilon},\\
    w_{pq}&=\max\!\big(M_{t,v}(p),M_{t,v}(q)\big),
\end{aligned}
\label{eq:grad_loss}
\end{equation}
so that pairs straddling the silhouette are kept. The structural term is
$\mathcal{L}_{\mathrm{ssim}}=1-\mathrm{SSIM}\big(\hat I_{t,v}|_{\mathcal{W}_{t,v}},\,
I_{t,v}|_{\mathcal{W}_{t,v}}\big)$, evaluated on the bounding box $\mathcal{W}_{t,v}$ of
$M_{t,v}$ padded by 8\,px.

\paragraph{Silhouette and covariance.}
The silhouette term is a whole-image L1 between the foreground opacity and the mask,
$\mathcal{L}_{\mathrm{sil}}=\frac{1}{|\Omega|}\sum_{p\in\Omega}
\big|\hat\alpha_{t,v}(p)-M_{t,v}(p)\big|$; its outside-mask half forbids opacity where no
liquid was observed. The covariance regularizer penalizes the total extent of the decoded
Gaussians,
$\mathcal{L}_{\mathrm{reg}}=\frac{1}{K|\mathcal{A}_t|}\sum_{i\in\mathcal{A}_t}\sum_{k}
\mathrm{tr}\big(\Sigma_{i,k,t}\big)$.

\paragraph{Separation and colour consistency.}
Following the pairwise separation regularization of~\cite{gao2025fluidnexus}, but restricted to the $K$ children of one carrier (the only pairs that can collapse, since a child is bounded to its own
carrier frame $A_{i,t}$), we penalize small distances between the bounded offsets
$o_{i,k}=\beta\tanh\big(g_o(f_i)_k\big)$ with a squared hinge penalty on distances below a margin.
\begin{equation}
    \mathcal{L}_{\mathrm{sep}}
    =
    \frac{1}{|\mathcal{A}_t|\binom{K}{2}}
    \sum_{i\in\mathcal{A}_t}\;\sum_{k<l}
    \max\!\big(0,\;\tau_{\mathrm{sep}}-\|o_{i,k}-o_{i,l}\|_2\big)^2.
    \label{eq:sep_loss}
\end{equation}

We set $\tau_{\mathrm{sep}}=0.6$ in carrier-local units, while the children are initialized approximately one unit apart.

Temporal consistency within a carrier is maintained by construction, since its feature $f_i$ is shared over the carrier's lifetime. The remaining flicker arises when carriers are removed and replaced, so
that independently decoded colours differ between nearby carriers. We therefore regularize colour consistency spatially. Carriers are
assigned to the voxel grid of $\phi_t$ by
$\mathrm{cell}(i)=\lfloor x_{i,t}/h \rfloor$ with voxel size $h$, and
for each voxel $j$ we compute the mean colour $\bar c_j$ over all
children of carriers in that voxel. Each decoded child colour
$c_{i,k}=g_c([f_i,\omega_{i,t,v}])_k$ is then pulled toward this local
mean,

\begin{equation}
    \mathcal{L}_{\mathrm{col}}
    =
    \frac{1}{3K|\mathcal{A}_t|}
    \sum_{i\in\mathcal{A}_t}\sum_{k}
    \big\|c_{i,k}-\bar c_{\mathrm{cell}(i)}\big\|_2^2 .
    \label{eq:col_loss}
\end{equation}

\paragraph{Total objective.}
The training objective of the main paper (Sec.~4.4) is the weighted sum:
\begin{equation}
\begin{aligned}
\mathcal{L}
={}& \lambda_{\mathrm{fg}}\mathcal{L}_{\mathrm{fg}}
   +\lambda_{\mathrm{grad}}\mathcal{L}_{\mathrm{grad}}
   +\lambda_{\mathrm{ssim}}\mathcal{L}_{\mathrm{ssim}}
   +\lambda_{\mathrm{sil}}\mathcal{L}_{\mathrm{sil}} \\
&{}+\lambda_{\mathrm{reg}}\mathcal{L}_{\mathrm{reg}}
   +\lambda_{\mathrm{sep}}\mathcal{L}_{\mathrm{sep}}
   +\lambda_{\mathrm{col}}\mathcal{L}_{\mathrm{col}}.
\label{eq:total_loss}
\end{aligned}
\end{equation}

\subsection{Implementation Details}
\label{sec:supp_impl}

\smallskip\noindent\textbf{Observation fields.}
The hull grid spans the union of the sequence-wide visual hulls with $10\%$ padding, 176 near-cubic cells along the longest axis. A voxel is occupied only if it projects inside the liquid masks of all five
training views, and $\phi_t$ is the signed distance to this region.
Eq.~(2) is solved on a 96-cell grid with $u_t$ parameterized by a $24^3$ control grid: 35 residual-descent steps of size $0.45$ within
a four-voxel band around the two interfaces. After each step the field is Gaussian-smoothed ($\sigma=0.6$ voxel), blended with $35\%$ of its divergence-free projection, and clipped to eight voxels per
frame.

\smallskip\noindent\textbf{Carriers.}
All distances are in transport-grid voxels. Carriers advance one frame by midpoint RK2, with the singular values of $F$ clamped to $[1/3,3]$, and are deactivated once $\phi_{t+1}>0.5$. Coverage is restored within the interior shell $-3\le\phi_t\le0$, where each occupied cell is assigned a target population $\bar n\in\{2,\dots,5\}$ from its proximity to the training-view silhouettes and the local curvature of $\phi_t$. New carriers are
drawn uniformly within the cell with $F=I$.

\smallskip\noindent\textbf{Decoder and optimization.}
Each carrier stores a 32-D latent feature and decodes $K=2$ child Gaussians. Offset, scale, and opacity are predicted by lightweight linear heads, while colour is decoded by a two-layer MLP with hidden width 64 and a separate RGB output for each child. 

We first train the Scaffold-GS background on frame~0 for 20k iterations and keep it fixed thereafter. The dynamic foreground is then optimized for 3k iterations using Adam with a learning rate of $2\times10^{-3}$, sampling one training frame--camera pair per iteration at the native crop resolution. We use fixed loss weights for all experiments:
$\lambda_{\mathrm{fg}}=0.5$, $\lambda_{\mathrm{grad}}=0.1$,
$\lambda_{\mathrm{ssim}}=0.2$, $\lambda_{\mathrm{sil}}=0.7$,
$\lambda_{\mathrm{reg}}=0.01$, $\lambda_{\mathrm{sep}}=0.1$, and
$\lambda_{\mathrm{col}}=1$.

\smallskip\noindent\textbf{Sub-frame decoding.}
Our carrier representation supports continuous-time queries between two observed frames. For an intermediate time $t+\tau$, $0<\tau<1$, we advect each carrier from frame $t$ using the fitted velocity field $u_t$. Because no observation is available at the intermediate time, we apply neither observation correction nor carrier resampling. The SDF used for decoding is instead obtained by linearly interpolating the two neighboring observed SDFs,
$\phi_{t+\tau}(x)=(1-\tau)\phi_t(x)+\tau\phi_{t+1}(x)$.
All learned carrier attributes remain unchanged during this interpolation.

For the interpolation experiment in Fig.~5, only the observed frames are used to construct the SDFs and motion fields; the withheld intermediate frames are used solely for evaluation. Thus, their images, masks, and geometry provide no input to the reconstruction.

\smallskip\noindent\textbf{Baselines.}
All baselines use their released implementations and recommended configurations, with the same data, camera split, crops, and evaluation protocol. The only changes are to data loading, extending Deformable-3DGS \cite{yang2023deformable3dgs} and SpacetimeGaussians \cite{Li_STG_2024_CVPR} to per-camera intrinsics and our two-view test split. The baselines don't supports\ mask supervision natively, and grafting such terms on would require per-method re-tuning that risks misrepresenting their behavior. The Per-frame variant of the main paper serves as the supervision-matched control. On the synthetic scenes every method, including ours, starts from a point cloud sampled from the mask-derived visual hull of the training views, never from ground-truth particles.

\section{Additional Comparison with Baselines}
\label{sec:supp_experiments}
 
\begin{figure*}[t]
\centering
\setlength{\tabcolsep}{0.000813\linewidth}
\newcommand{\hcell}[1]{\parbox{0.123577\linewidth}{\centering\small #1}}
\begin{tabular}{@{}cccccccc@{}}
\hcell{reference} & \hcell{captured $t$} & \hcell{D3G} &
\hcell{STG} & \hcell{4DSG} & \hcell{Ours} &
\hcell{ground truth} & \hcell{captured $t\!+\!1$} \\[1pt]
\end{tabular}\\[1pt]
\includegraphics[width=\linewidth]{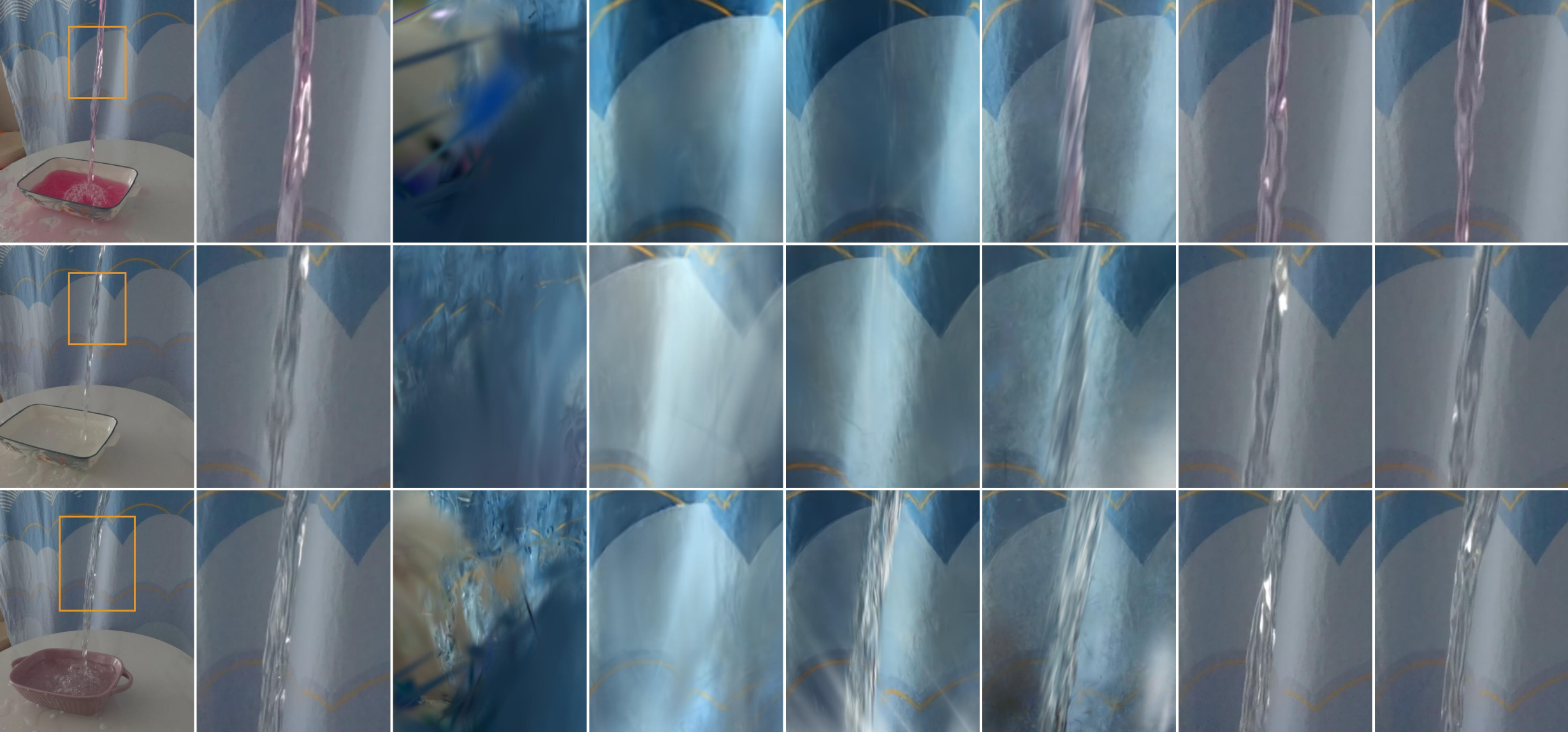}
\caption{\textbf{Rendering between captured frames, all baselines.} The instant in the four middle columns was never trained on; the outer columns are the captures that bracket it and the amber box in the reference marks the crop. Neither the deformation-field nor the polynomial-motion baselines place the falling stream where the withheld capture puts it.}
\label{fig:suppl-interp}
\end{figure*}

\begin{table}[t]\centering\small
\begin{tabular}{lcccc}
\toprule
& PSNR $\uparrow$ & SSIM $\uparrow$ & LPIPS $\downarrow$ & $|\Delta\mathrm{Jitter}|\downarrow$ \\
\midrule
D3G~\cite{yang2023deformable3dgs}   & 13.94 & 0.930 & 0.121 & 0.065 \\
STG~\cite{Li_STG_2024_CVPR}         & 17.92 & 0.947 & 0.092 & 0.061 \\
4DSG~\cite{cho20264d}             & 19.16 & \textbf{0.949} & 0.077 & 0.046 \\
Ours              & \textbf{19.92} & 0.944 & \textbf{0.066} & \textbf{0.024} \\
\bottomrule
\end{tabular}
\caption{\textbf{Hidden-frame interpolation}, mean over $10$ scenes on held-out cameras.}
\label{tab:suppl-interp}
\end{table}

\subsection{Additional Temporal Interpolation Comparison.}
We extend the comparison of the main paper to the two remaining baselines under the same
protocol. Deformable-3DGS conditions on a normalised time and SpacetimeGaussians give each
Gaussian a polynomial trajectory; both can therefore be queried between the frames they were
fitted to, and both interpolate appearance rather than transport material.
\cref{tab:suppl-interp} reports the means over all $10$ scenes and \cref{fig:suppl-interp}
the corresponding renderings. Ours is best on every metric except SSIM, and best on $10/10$
scenes for both LPIPS and $|\Delta\mathrm{Jitter}|$, where its temporal error is less than half
that of the closest baseline. 4D-Scaffold-GS retains a small SSIM advantage but the motion it reproduces is not the captured one. The deformation field of Deformable-3DGS overfits severely with five training views, which the qualitative comparison makes plain.

\subsection{Additional Comparison on Real Capture}
\begin{figure}[t]
  \centering
  \includegraphics[width=\linewidth]{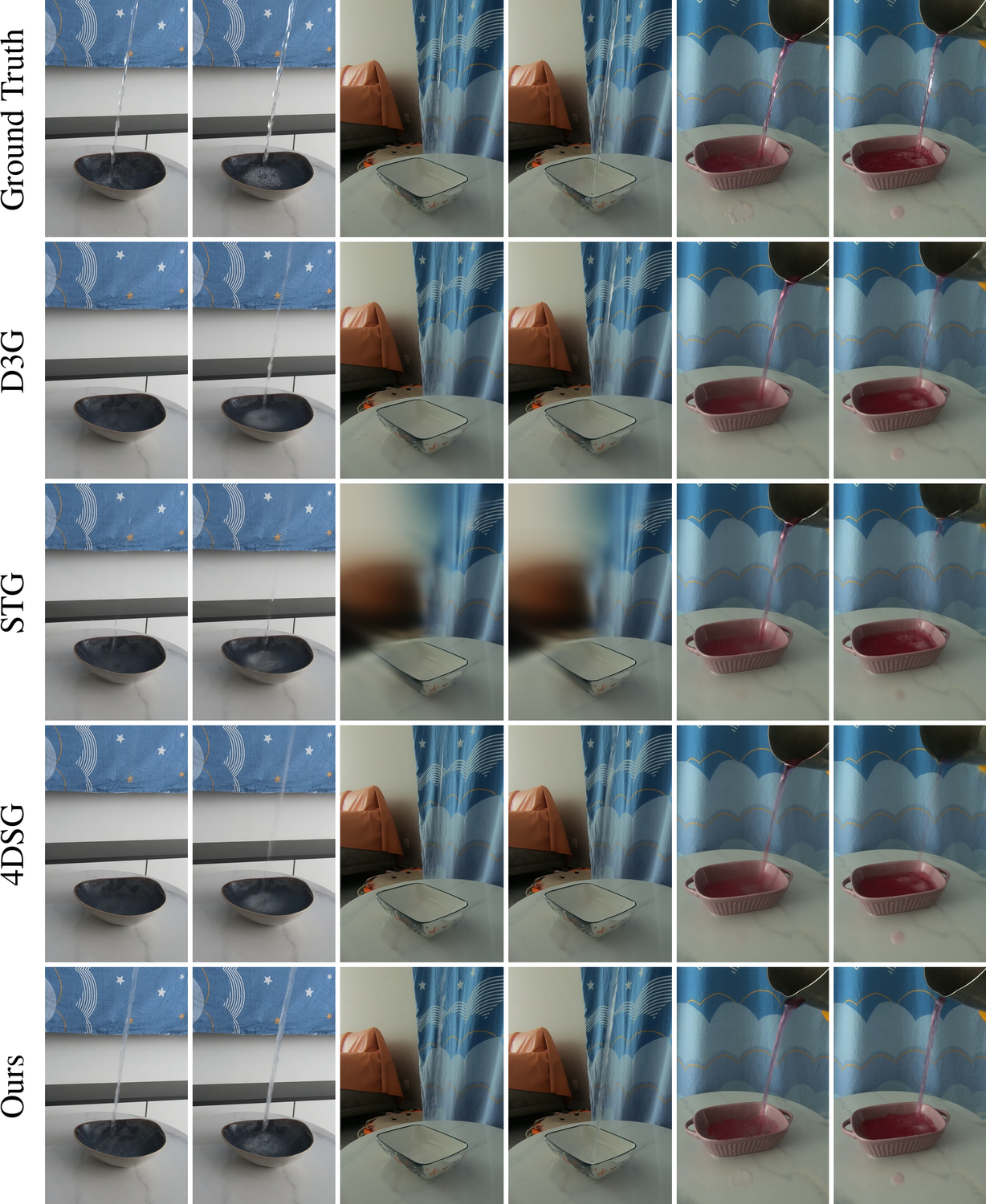}
  \caption{\textbf{Training views on real captures}, cropped to the fluid. One scene per column (two frames each), one method per row. Even on held-in views the baselines thin out or drop the falling stream, whereas our reconstruction keeps it continuous down to the pool.}
  \label{fig:crop_showcase_train_bymethod}
\end{figure}

\begin{figure}[t]
  \centering
  \includegraphics[width=\linewidth]{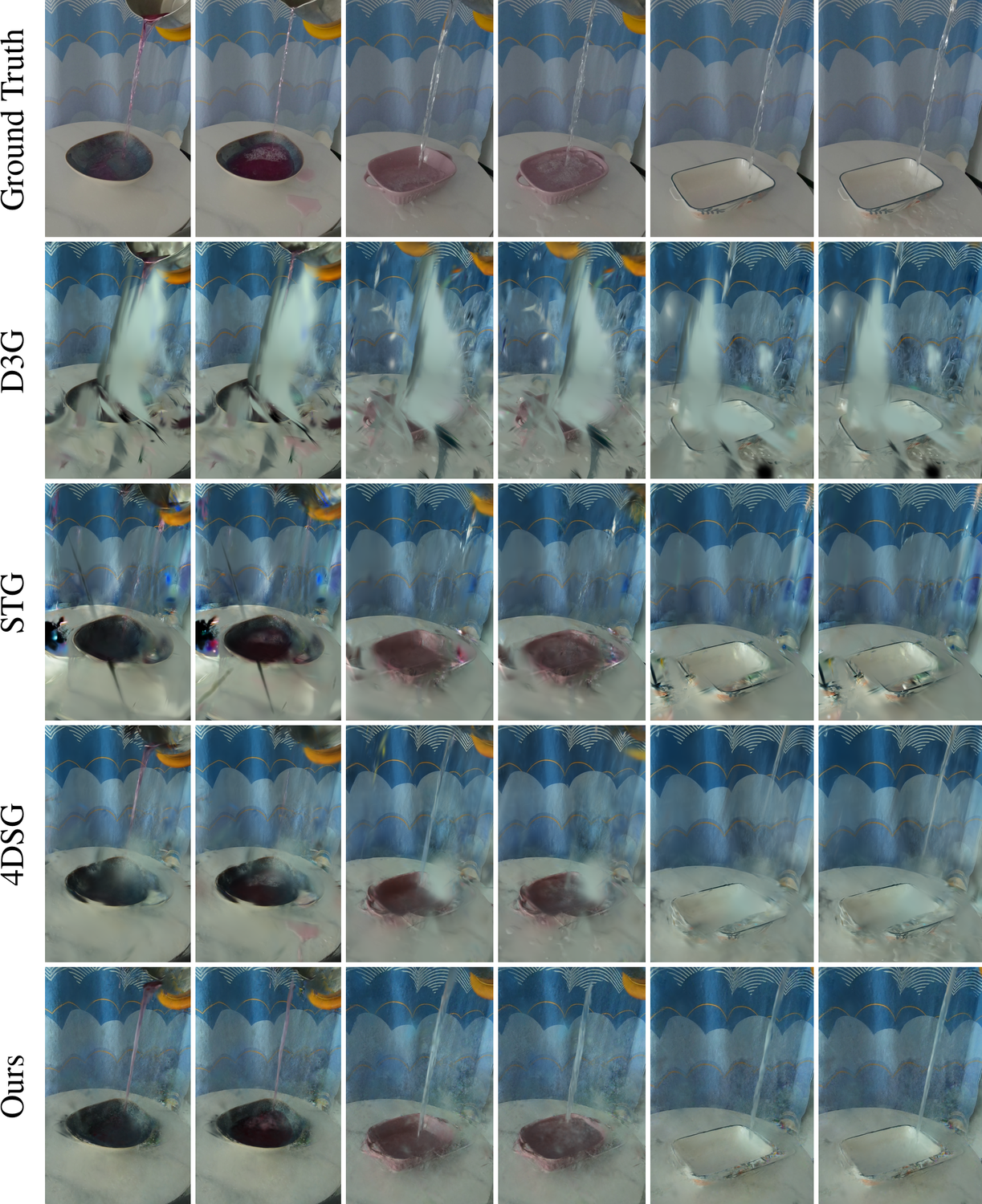}
  \caption{\textbf{Held-out views on real captures}, cropped to the fluid. Same layout as the training-view figure. On novel views the baselines lose the stream and smear the container, while ours keeps the stream and the pool readable.}
  \label{fig:crop_showcase_test_bymethod}
\end{figure}

Our carriers show substantially more uniform spatial coverage than the baselines: even after normalizing by mean density, $\sigma_D/\mu_D$ is $0.198$ against $0.429$--$0.686$ (\cref{tab:supp_comp}). Since the normalized form controls for primitive count, the difference reflects how the primitives are placed rather than how many there are: observation-guided resampling redistributes carriers over the reconstructed liquid each frame, whereas unconstrained Gaussian optimization concentrates primitives
where the appearance objective favours them. Rendering speed and storage are competitive: SpacetimeGaussians renders fastest, since its per-Gaussian attributes are evaluated in closed form from stored polynomial and temporal RBF parameters, while our runtime remains
comparable to 4D-Scaffold-GS despite the per-frame carrier update, and the shared decoder heads keep the model compact.
We provide additional qualitative comparision in \cref{fig:crop_showcase_train_bymethod} and \cref{fig:crop_showcase_test_bymethod}.

\begin{table}[t]
  \centering
  \caption{\textbf{Additional metrics for baselines.} Rendering speed and storage for all methods.}
  \label{tab:supp_comp}
  \small
  \begin{tabular}{lcccc}
    \toprule
    Metric & D3G & STG & 4DSG & Ours \\
    \midrule
    FPS$\uparrow$          & 21.6 & \textbf{208.6} & 96.7 & 92.6 \\
    Storage (MB)$\downarrow$ & 74.0 & 54.9 & \textbf{40.4} & 42.6 \\
     $\sigma_D/\mu_D\downarrow$ & 0.686 & 0.429 & 0.642 & \textbf{0.198} \\
    \bottomrule
  \end{tabular}
\end{table}

\subsection{Additional NeuroFluid Evaluation}

We extend the evaluation of the main paper to the NeuroFluid
\emph{WaterCube} scene under the same protocol
(\cref{tab:supp_neurofluid}, \cref{fig:watercube_showcase}). On this
smooth-deformation scene the image metrics separate the methods by
fractions of a dB, while our temporal error is several times lower
than the best baseline's, the pattern the main paper predicts: a fit
that explains held-out views need not reproduce the captured motion.
Qualitatively, STG scatters debris off the surface and D3G and 4DSG
smooth away the rim droplets, while ours preserves the rim and splash
structure.
 
\begin{table}[t]
  \centering
  \caption{\textbf{Additional NeuroFluid scenes.} Foreground novel-view
  synthesis on scene WaterCube under the protocol of the main paper.}
  \label{tab:supp_neurofluid}
  \small
  \begin{tabular}{lcccc}
    \toprule
    Method & PSNR$\uparrow$ & SSIM$\uparrow$ & LPIPS$\downarrow$ &
    $|\Delta\mathrm{Jitter}|\downarrow$ \\
    \midrule
    D3G    & 31.89 & 0.9325 & 0.1861 & 0.0065 \\
   STG & 28.13 & 0.8633 & 0.2778 & 0.0138 \\
    4DSG     & 31.48 & 0.9340 & 0.1725 & 0.0057 \\
    Ours               & \textbf{31.93} & \textbf{0.9355} &
                         \textbf{0.1721} & \textbf{0.0012} \\
    \bottomrule
  \end{tabular}
\end{table}

\begin{figure}[t]
\centering
\setlength{\tabcolsep}{0.001200\linewidth}
\newcommand{\scell}[1]{\parbox{0.248199\linewidth}{\centering\small #1}}
\begin{tabular}{@{}cccc@{}}
\scell{Reconstruction} & \scell{Amber Dye} & \scell{Brown Dye} & \scell{Green Dye} \\[1pt]
\end{tabular}\\[1pt]
\includegraphics[width=\linewidth]{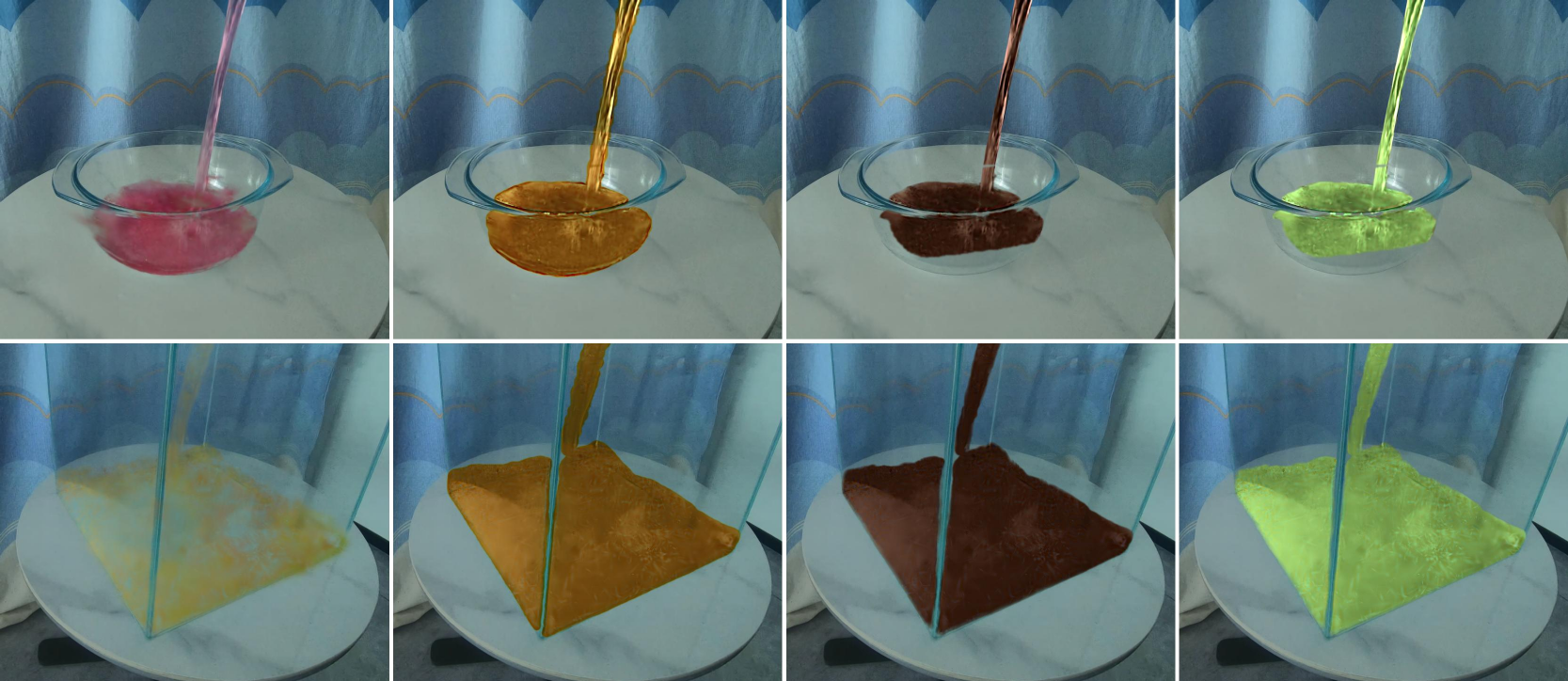}
\caption{\textbf{Further dyes and vessels.} Each row is a single reconstruction re-rendered with a different absorption spectrum. The vessel, the table and the background are untouched, and both containers are transparent, so the wall stays visible in front of the liquid while the pool deepens in colour with accumulated volume.}
\label{fig:suppl-style}
\end{figure}

\subsection{Additional Style Transfer Results.}
A dye is an absorbing dielectric: it attenuates the refracted background along the optical path, $T_c=\exp(-\sigma_c\,\ell)$ per colour channel $c$, with $\ell$ the path length through the liquid and $\sigma_c$ its absorption coefficient~\cite{max1995optical}. A multiple-scattering term lets a thick pool converge to a saturated colour, and the interface is
composited with a Fresnel term~\cite{schlick1994fresnel}. Both quantities the model needs, the optical path length and the interface normal, are read directly from the reconstructed surface and carriers, so a new material is a change of coefficients rather than a re-fit and costs a few seconds. The recorded stream structure is preserved throughout: the ripples and highlights of \cref{fig:suppl-style} are the captured ones, modulated by the new absorption rather than replaced.

\section{Extended Ablation Studies}
\label{sec:supp_ablations}
 
\subsection{Mask Robustness}
We assess the dependence on manual annotation by training the full
model on raw SAM3 masks (Fig.~\ref{fig:supp_masks}) instead of the
refined ones. Removing the refinement pass costs silhouette accuracy,
while the photometric and physical scores degrade only slightly
(Tab.~\ref{tab:supp_maskrobust}). The dependence on manual annotation
is therefore concentrated in the recovered geometry rather than in
the rendering metrics.
 
\begin{table}[t]
  \centering
  \caption{\textbf{Mask robustness.} Full model trained on raw SAM3
  masks and on manually refined masks.}
  \label{tab:supp_maskrobust}
  \small
  \begin{tabular}{lcccc}
  \toprule
  Masks & mIoU@0.5$\uparrow$ & PSNR$\uparrow$ & LPIPS$\downarrow$ &
  $\sigma_D\downarrow$ \\
  \midrule
  Raw SAM3 & 0.665 & 19.60 & 0.0858 & 258.5 \\
  Refined  & \textbf{0.770} & \textbf{20.31} & \textbf{0.0764} & \textbf{234.2} \\
  \bottomrule
\end{tabular}
\end{table}

\subsection{Number of Training Views}
We reduce the number of training views from five to three and measure
how reconstruction degrades as the visual hull loosens, delineating
the operating range of the benchmark (Tab.~\ref{tab:supp_views}).
Every mask-derived stage---SDF, flow fields, background scaffold, and
foreground---is rebuilt from the reduced set, so each arm is a
genuine $N$-camera capture rather than a five-camera pipeline scored
on fewer views; the held-out cameras are unchanged, and the nested
subsets keep both as interpolation views. Silhouette accuracy holds
at four views and drops at three, as fewer carving planes inflate the
hull. PSNR falls earlier, but much of that drop comes from the frozen
background, which is also refit from the reduced views.
\begin{table}[t]
\centering
\caption{\textbf{Number of training views.} Full model trained with
five, four, and three of the seven calibrated cameras on
\texttt{bowl\_001}--\texttt{bowl\_010}.}
\label{tab:supp_views}
\small
\setlength{\tabcolsep}{4pt}
\begin{tabular}{lccc}
\toprule
Views & mIoU@0.5$\uparrow$ & PSNR$\uparrow$ &
LPIPS$\downarrow$ \\
\midrule
5 (full) & \textbf{0.770} & \textbf{20.31} & \textbf{0.0764} \\
4        & 0.753 & 18.52 & 0.0840 \\
3        & 0.680 & 18.34 & 0.0845 \\
\bottomrule
\end{tabular}
\end{table}

\subsection{Hyperparameter Sensitivity}

Tab.~\ref{tab:supp_sens} sweeps the main hyperparameters one at a
time. Grid resolution has the clearest effect: a coarser grid lacks
the spatial resolution to represent the liquid geometry and
transport, giving less uniform carrier distributions and larger
energy deviations, while a finer grid improves the physical measures
only marginally at roughly twice the decoded Gaussians and storage.
The number of children behaves similarly, with $K{=}1$ reducing
capacity and larger $K$ adding rendering cost without appearance
gain, and raising the minimum carrier density adds primitives
throughout the liquid for negligible benefit. The default setting is
therefore not a narrow optimum but the point where further capacity
no longer benefits the model.

\begin{table}[t]
  \centering
  \caption{\textbf{Hyperparameter sensitivity.}
  Single-knob variants of the full model (defaults: grid $96$, $K=2$, $p_{\min}=2$); ten sequences, held-out views, foreground
  metrics as in the main paper. $\sigma_D/\mu_D$ is count-matched at $16$k, and the $K$ variants share the full model's pool. Gauss./fr.\
  counts the Gaussians decoded per frame, and Storage is the dynamic
  model's checkpoint (training peaks at ${\approx}5$\,GiB for every
  variant).}
  \label{tab:supp_sens}
  \footnotesize
  \setlength{\tabcolsep}{3pt}
  \begin{tabular}{lcccccc}
    \toprule
    Variant & PSNR$\uparrow$ & LPIPS$\downarrow$ & $\sigma_D/\mu_D\downarrow$ &
    $\Delta E\downarrow$ & Gauss./fr. & Storage \\
    \midrule
    Grid~64            & 19.81 & 0.0808 & 0.417  & 0.426  & 30k & 6\,MB \\
    Grid~128            & 19.91 & \textbf{0.0752} & \textbf{0.158} & \textbf{0.366} & 179k & 39\,MB \\
    $K{=}1$             & 19.83 & 0.0816 & 0.162 & 0.374 & 57k & 17\,MB \\
    $K{=}4$             & 19.91 & 0.0756 & 0.162 & 0.374 & 171k & 17\,MB \\
    $p_{\min}{=}4$ & 19.95 & 0.0758 & 0.163 & 0.373 & 140k & 27\,MB \\
    $p_{\min}{=}1$ & 19.87 & 0.0789 &0.280 & 0.385 & 78k & 15\,MB \\
    \midrule
    \textbf{Full model} & \textbf{20.31} & 0.0764 & 0.162 & 0.374 & 85k & 17\,MB \\
    \bottomrule
  \end{tabular}
\end{table}

\begin{table}[t]
\centering
\caption{\textbf{One-step carrier transport under matched
initialization.} All predictors start from the same observed state.
Validity is the fraction of carriers still inside the observed liquid
after one step, measured by sampling $\phi_{t+1}$ at each carrier
($\phi<0.5$\,voxel); $E_{\mathrm{out}}$ is the mean escape distance
$\max(\phi,0)$ in voxels. Lower block: solver variants on one
\textsc{Bowl} scene.}
\label{tab:flip_onestep}
\footnotesize
\setlength{\tabcolsep}{4pt}
\renewcommand{\arraystretch}{0.95}
\begin{tabular}{l cc}
\toprule
Predictor & Validity$\uparrow$ & $E_{\mathrm{out}}\downarrow$ \\
\midrule
Ours                          & \textbf{0.956} & \textbf{0.06} \\
No motion                     & 0.928 & 0.11 \\
FLIP                          & 0.840 & 0.25 \\
\midrule
FLIP, $2\times$ substeps      & 0.868 & 0.19 \\
FLIP, $2\times$ press.\ iters & 0.858 & 0.21 \\
FLIP, $2\times$ grid          & 0.846 & 0.22 \\
\bottomrule
\end{tabular}
\end{table}

\subsection{One-Step Solver Comparison}
The main paper reports that replacing our fitted transport with an
open-loop FLIP solver degrades reconstruction. To test whether this
stems from the solver rather than from its interaction with the
reconstruction, we isolate a single transport step from an aligned
observed state. For every $t\!\rightarrow\!t{+}1$ all predictors
start from the same $\phi_t$ and $u_t$; FLIP additionally receives
particle and grid velocities sampled from $u_t$, the registered
container geometry, and gravity in the metric scale of the capture.

Our transport keeps $95.6\%$ of carriers inside the observed liquid after one step, against $84.0\%$ for FLIP. Leaving the carriers in
place scores $92.8\%$, since over one frame the interface moves little relative to the voxel scale. FLIP falls below this because its initial and boundary conditions are incomplete rather than incorrect: the interior velocity field is unobservable and is therefore seeded from
the surface-derived $u_t$, and the pour has no inflow specification, so the simulated volume stays fixed while the observed volume grows.
Under these conditions the step is dominated by free fall and carries carriers past the observed interface. Doubling substeps, pressure iterations, or grid resolution does not close the gap, so the discrepancy is not one of discretization or convergence. Supplying the missing conditions would require knowing the inflow rate and the interior state, neither of which the images reveal.

\begin{figure}[t]
\centering
\setlength{\tabcolsep}{0.001200\linewidth}
\newcommand{\scell}[1]{\parbox{0.18\linewidth}{\centering\small #1}}
\begin{tabular}{@{}ccccc@{}}
\scell{Ground Truth} & \scell{D3G} & \scell{STG} & \scell{4DSG} & \scell{Ours} \\[1pt]
\end{tabular}\\[1pt]
  \includegraphics[width=\linewidth]{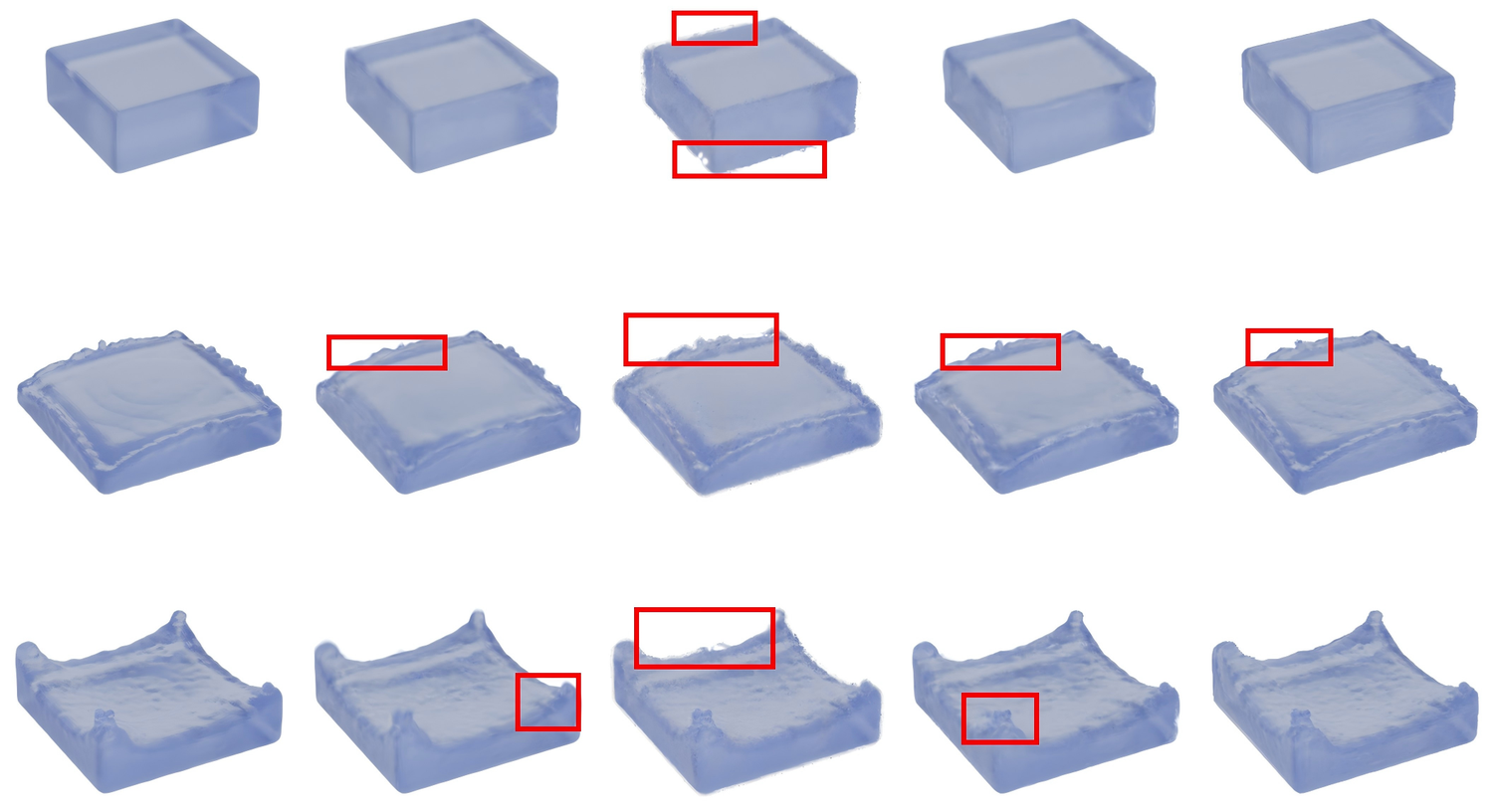}
  \caption{Novel view synthesis on NeuroFluid \emph{WaterCube} (held-out view, $t=6,18,40$).}
  \label{fig:watercube_showcase}
\end{figure}

\end{document}